\documentclass[10pt,twocolumn,letterpaper]{article}

\usepackage[pagenumbers]{iccv} 

\definecolor{iccvblue}{rgb}{0.21,0.49,0.74}
\usepackage[pagebackref,breaklinks,colorlinks,allcolors=iccvblue]{hyperref}

\usepackage{array}
\newcolumntype{C}[1]{>{\centering\let\newline\\\arraybackslash\hspace{0pt}}m{#1}}

\usepackage{color, colortbl}
\definecolor{Gray}{gray}{0.85}
\usepackage{lipsum}
\usepackage{bbm}
\usepackage{verbatim}
\usepackage{multirow}
\usepackage{amsmath}

\usepackage{multirow}
\usepackage{algpseudocode}
\usepackage{algorithm}
\usepackage{algorithmicx}
\usepackage{setspace}
\usepackage{graphicx}
\usepackage{adjustbox}
\usepackage{comment}
\usepackage[normalem]{ulem}
\usepackage{bbm}
\usepackage{helvet}
\usepackage{tikz}
\usetikzlibrary{fadings,shapes.arrows,shadows}   
\tikzset{arrowstyle/.style={draw= black,single arrow,minimum height=#1, single arrow,
single arrow head extend=.4cm,align=center }}
\usepackage{amsthm}
\usepackage{amssymb}
\usepackage{wrapfig}

\usepackage{subcaption}
\usepackage{cellspace}
\usepackage{booktabs} 
\usepackage{caption} 
\usepackage{bm}
\usepackage{dsfont}

\usepackage{xcolor}

\theoremstyle{plain}

\theoremstyle{definition}

\theoremstyle{remark}

\def\paperID{3719} 
\def\confName{ICCV}
\def\confYear{2025}

\title{Continual Multiple Instance Learning with Enhanced Localization \\ for Histopathological Whole Slide Image Analysis}

\author{Byung Hyun Lee\textsuperscript{1},\,\,\, Wongi Jeong\textsuperscript{1},\,\,\, 
Woojae Han\textsuperscript{2},\,\,\, Kyoungbun Lee\textsuperscript{4},\,\,\,
Se Young Chun\textsuperscript{1,}\textsuperscript{2,}\textsuperscript{3,}\textsuperscript{$\dagger$} \\
\textsuperscript{1}Department of ECE, \textsuperscript{2}IPAI, \textsuperscript{3}INMC, Seoul National University\\
\textsuperscript{4}Department of Pathology, College of Medicine, Seoul National University\\
{\tt \small \{ldlqudgus756, wg7139, wjhan24, sychun\}@snu.ac.kr}\\
}

\begin{document}
\maketitle
\renewcommand{\thefootnote}{}
\footnotetext{$\dagger$\ Corresponding author.}

\begin{abstract} 
Multiple instance learning (MIL) significantly reduced annotation costs via bag-level weak labels for large-scale images, such as histopathological whole slide images (WSIs). However, its adaptability to continual tasks with minimal forgetting has been rarely explored, especially on instance classification for localization. Weakly incremental learning for semantic segmentation has been studied for continual localization, but it focused on natural images, leveraging global relationships among hundreds of small patches (e.g., $16 \times 16$) using pre-trained models. This approach seems infeasible for MIL localization due to enormous amounts ($\sim 10^5$) of large patches (e.g., $256 \times 256$) and no available global relationships such as cancer cells. To address these challenges, we propose Continual Multiple Instance Learning with Enhanced Localization (CoMEL), an MIL framework  for both localization and adaptability with minimal forgetting. CoMEL consists of (1) Grouped Double Attention Transformer (GDAT) for efficient instance encoding, (2) Bag Prototypes-based Pseudo-Labeling (BPPL) for reliable instance pseudo-labeling, and (3) Orthogonal Weighted Low-Rank Adaptation (OWLoRA) to mitigate forgetting in both bag and instance classification. Extensive experiments on three public WSI datasets demonstrate superior performance of CoMEL, outperforming the prior arts by up to $11.00\%$ in bag-level accuracy and up to $23.4\%$ in localization accuracy under the continual MIL setup.
\end{abstract}

\section{Introduction}

Multiple instance learning (MIL) \cite{ilse2018attention, li2021dual, li2021dt, chen2022scaling, zheng2022graph, zhang2022dtfd, liu2024advmil, zhu2025dgr, tang2024feature, liu2023multiple, wang2024rethinking, castrosm} is a promising approach to mitigate high annotation costs by utilizing only bag-level (or slide-level) weak labels for large-scale images like histopathological whole slide images (WSIs). To effectively utilize the weak labels, MIL learns to predict the slide-level class by treating each slide as a bag of multiple instances (or patches).

For practical utility, it is required to learn continuously arriving tasks while retaining previous knowledge without forgetting.
As a real-world scenario, for example, about 1,000 WSIs are generated every working day in a 1700-bed-sized tertiary hospital, annually resulting in 500 terabytes of data.
Since this rapidly growing data volume could easily exceed the storage capacity, MIL models need to continually adapt to new 
conditions like organs and tumor subtypes. 

\begin{figure}[t]
\begin{center}
\centerline{\includegraphics[width=1.0\columnwidth]{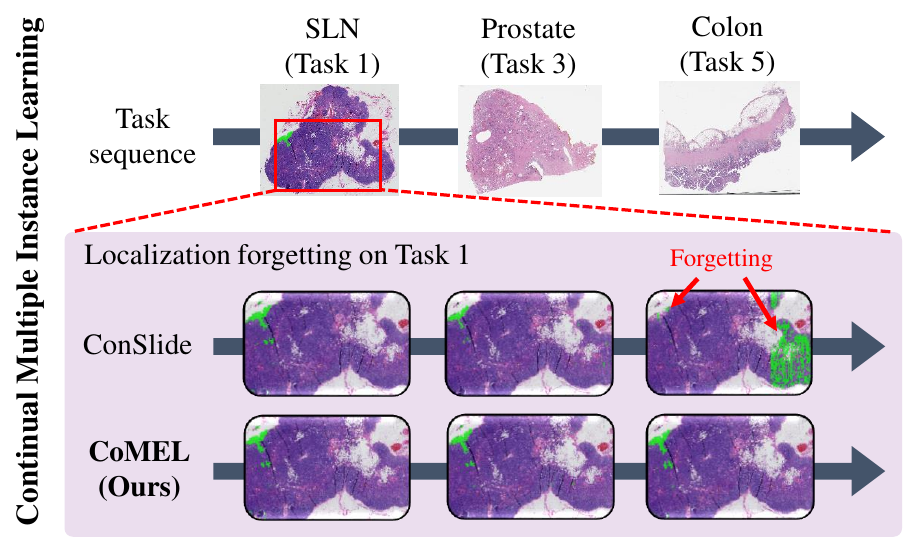}}
 \vskip -0.1in

\caption{We propose Continual Multiple instance learning with Enhanced Localization (CoMEL). Compared to the baseline \cite{huang2023conslide}, it alleviates forgetting of localization under continual MIL setup.}

\label{fig:overall_framework}
\vskip -0.4in
\end{center}
\end{figure}


However, its adaptability to continually varying conditions with minimal forgetting has rarely been  explored, especially on instance classification for localization.
A few studies \cite{huang2023conslide, gou2024queryable} integrated continual learning \cite{aljundi2018memory, lee2023online, qiaolearn} into MIL to efficiently adapt to sequential tasks while mitigating forgetting. They achieved excellent performance in continual bag classification, but their evaluation remains underexplored on instance classification for localization, an equally important aspect of MIL as bag classification.

Meanwhile, weakly incremental learning for semantic segmentation \cite{cermelli2022incremental, yu2023foundation, hsieh2023class, liu2024learning, si2024tendency} has been studied for continual localization. By utilizing rehearsal-based approaches \cite{aljundi2018memory, buzzega2020dark, lee2024doubly, chaudhry2021using} and dense pseudo-labels on data from previous tasks, they effectively mitigated forgetting of the localization. However, it focused on natural images, leveraging global relationships among small patches (e.g., $16 \times 16$) by pre-trained models. This approach is infeasible for MIL localization due to enormous amount ($\sim 10^5$) of large patches (e.g., $256 \times 256$), separately extracted from pre-trained models, without clear global relationships among cancer tissues.


To address these challenges, we propose an MIL framework dubbed Continual Multiple instance learning with Enhanced Localization (CoMEL), illustrated in \cref{fig:overall_framework}. In CoMEL, we aim to effectively train on both bag and instance classification with minimal forgetting under the continual MIL setup. 
To improve instance classification for localization, we first utilize an efficient attention mechanism called Grouped Double Attention Transformer (GDAT), by grouping instances into a small number of tokens and applying two attentions to them.
Then, we propose Bag Prototypes-based Pseudo-Labeling (BPPL), which assigns pseudo-labels to instances by comparing with bag prototypes and filtering them using thresholds adaptive to the model’s status on bag and instance classification. 
For continual MIL, we propose Orthogonal Weighted Low-Rank Adaptation (OWLoRA), which effectively mitigates forgetting by introducing new learnable bases for weights in our MIL model, orthogonal to previously learned bases.  
Based on the evaluations on 3 WSI datasets like CAMELYON-16 \cite{bejnordi2017diagnostic}, Pathology AI Platform (PAIP) \cite{kim2021paip,wisepaip}, and TCGA \cite{weinstein2013cancer}, our CoMEL outperformed prior arts on both continual bag and instance classification and achieved minimal forgetting on previous tasks. Our contributions are summarized:
\begin{itemize}
    \item We propose Continual Multiple instance learning with Enhanced Localization (CoMEL), an MIL framework focusing on both bag and instance classification under continually varying tasks with minimal forgetting.

    \item As main components, we propose Grouped Double Attention Transformer (GDAT) and Bag Prototypes-based Pseudo-Labeling (BPPL) for enhanced localization, and Orthogonal Weighted Low-Rank Adaptation (OWLoRA) for continual bag and instance classification.

    \item Extensive experiments on public WSI datasets demonstrate superior performance of CoMEL, achieving significantly reduced forgetting and outperforming prior arts by up to $11.00\%$ in bag-level accuracy and up to $23.4\%$ in localization accuracy (IoU) under the continual MIL setup.

\end{itemize}


\section{Related Works} \label{sec:related_works}
\textbf{Multiple Instance Learning (MIL)} \cite{ilse2018attention, li2021dual,tang2023multiple,chen2025cdp} is a widely utilized paradigm for analysis of computational pathology.
Traditional MIL methods utilized straightforward aggregation techniques such as max-/mean-pooling \cite{feng2017deep, zhu2017deep}. To improve the performance, attention-based aggregators \cite{ilse2018attention, lu2021data} were proposed to integrate attention mechanism and aggregate instances into a bag representation. It also effectively identified critical instances for bag classification. 
Different modifications were proposed based on the attention mechanism by the use of clustering layers \cite{lu2021data}, pseudo-bags \cite{zhang2022dtfd}, and similarity measures to refine attention \cite{li2021dual}. Especially, TransMIL \cite{shao2021transmil} proposed a transformer-based MIL to incorporate positional dependencies between the instances in a bag, extended by deformable transformer \cite{li2021dt}, hierarchical attention \cite{chen2022scaling}, or spatial-encoding transformer \cite{zhao2022setmil}.
$\text{R}^{2}$T-MIL suggested an efficient self-attention for re-embedding the instance features to refine and capture fine-grained local features.

\textbf{Continual Learning (CL)} Continual learning (CL) aims to retain previously learned knowledge while progressively learning new data. 
The landscape of CL includes three primary categories: class-incremental, task-incremental, and domain-incremental learning~\cite{de2021continual, van2018three}. 
Existing CL strategies are usually categorized into several groups: regularization-based \cite{kirkpatrick2017overcoming, zenke2017continual, li2017learning, chaudhry2018riemannian, wang2021afec}, architecture-based \cite{mallya2018packnet, serra2018overcoming, mallya2018piggyback, wang2023attriclip}, and rehearsal-based approaches \cite{lopez2017gradient,rolnick2019experience,buzzega2020dark,aljundi2018memory,lee2023online,lee2024doubly}. Recently, rehearsal-free CL approaches with Parameter-Efficient Tuning (PEFT) \cite{jia2022visual, hulora} has achieved competitive performance with or even surpassing state-of-the-art (SoTA) rehearsal-based methods.  
Prompt-tuning-based approaches \cite{wang2022learning, wang2022dualprompt, smith2023coda,   wang2023hierarchical, yang2024rcs, lemixture} train a small set of task-adaptive learnable prompts, thereby improving adaptability to new tasks mitigating forgetting. However, they require to add new learnable prompts for new tasks leading to continuous expansion of the model or either impose scalability limitations on the number of tasks when using a fixed prompt pool. Without altering the model structure, Low-Rank Adaptation (LoRA)-based approaches \cite{wang2023orthogonal, liang2024inflora, qiaolearn} decomposes linear projections in the model and regularize the orthogonality of components learned for each task. They verified their effectiveness to reduce task interference, significantly improving both forgetting and adaptability.

\textbf{Localization and CL for MIL.} \,\,
Despite the success of MIL, two factors constrain their practical applicability: (1) instance localization and (2) adaptability to sequentially changing environments. Recent works explored enhancing localization capability by fine-tuning MIL backbones using pseudo-labels \cite{liu2023multiple,li2023task,wang2024rethinking,shi2020loss,qu2022bi,liu2024weakly,yufei2022bayes,javed2022additive} or by regularizing the spatial smoothness of attentions \cite{castrosm}. Additionally, continual MIL frameworks have been introduced to improve adaptability to continuously varying conditions \cite{bandi2023continual, huang2023conslide, gou2024queryable}. ConSlide \cite{huang2023conslide} leveraged instance-level memory to efficiently retain knowledge of previous tasks with reduced resource consumption. However, since storing samples from prior tasks could raise privacy concerns, QPMIL-VL \cite{gou2024queryable} designed a prompt-tuning-based \cite{jia2022visual} continual MIL approach to prevent forgetting without memory.

\begin{figure}[t]
\begin{center}
\centerline{\includegraphics[width=1.0\columnwidth]{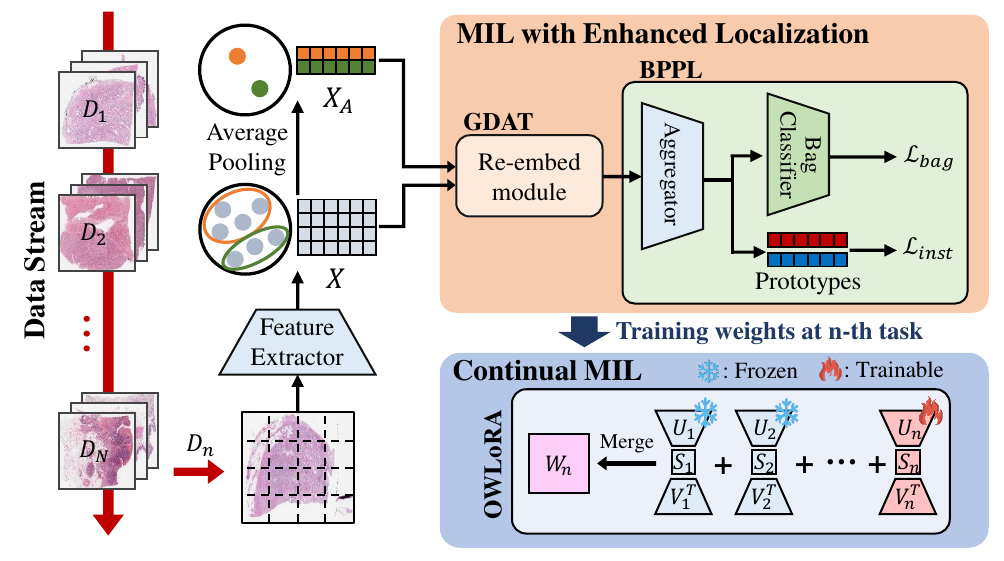}}
 \vskip -0.05in
\caption{Overview of Continual Multiple instance learning with Enhanced Localization (CoMEL). It aims to alleviate the forgetting of both bag and instance classification on previous tasks under the MIL setup, while effectively learning new tasks. It consists of three key components: Grouped Double Attention Transformer (GDAT), Bag Prototypes-based Pseudo-Labeling (BPPL), and Orthogonal Weighted Low-Rank Adapatation (OWLoRA).
}
\label{fig:CoMEL_overall_framework}
\vskip -0.4in
\end{center}
\end{figure}

\section{Continual MIL with Enhanced Localization}


We describe our framework dubbed Continual Multiple Instance Learning with Enhanced Localization (CoMEL), illustrated in \cref{fig:CoMEL_overall_framework}. CoMEL focuses on three challenging aspects: scalability, localization, and adaptability without forgetting.
For scalability, we first utilize an efficient attention mechanism in MIL to refine instance features (\cref{sec:cdat}). Then, we propose a pseudo-labeling scheme adaptive to the model's status for bag and instance classification to enhance localization through bag prototypes for localization (\cref{sec:ppl}). For continual MIL on both bag and instance classification, we propose a rehearsal-free approach based on LoRA by regularizing weight orthogonality (\cref{sec:owlora}).

\begin{figure*}[t]
\begin{center}

\centerline{\includegraphics[width=1.0\textwidth]{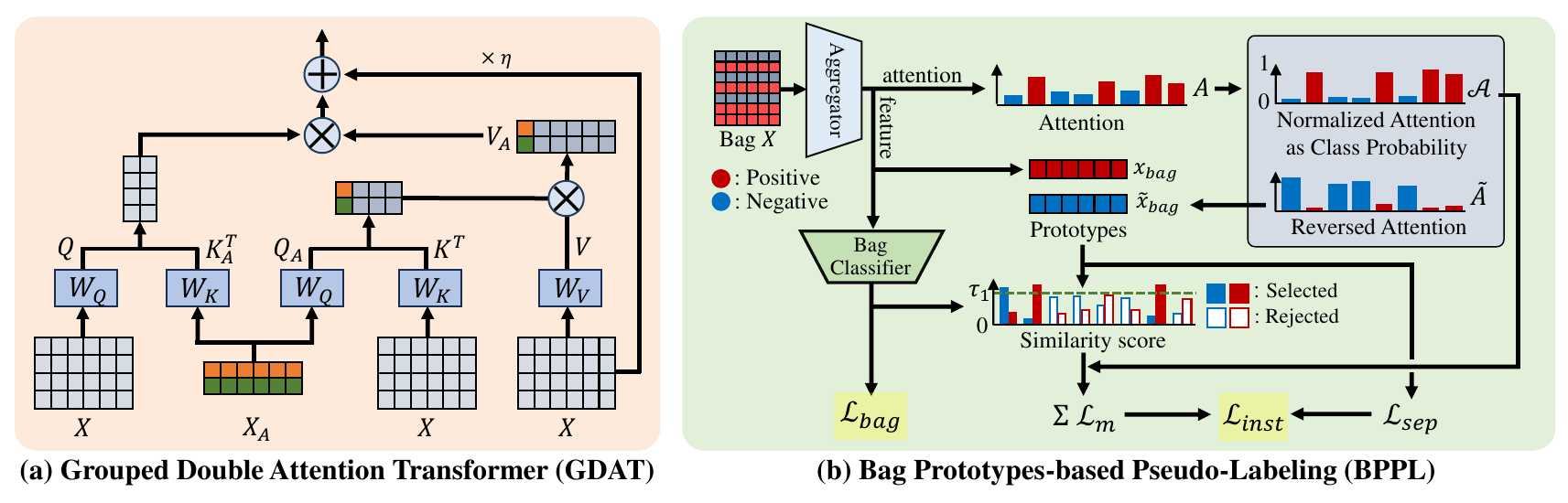}}
\vskip -0.1in
\caption{Components of (a) Grouped Double Attention Transformer (GDAT) and (b) Bag Prototypes-based Pseudo-Labeling (BPPL) for enhanced localization. GDAT utilizes two sequential efficient attention with small number of grouped tokens by averaging instances in same region into one token. For BPPL, we normalize the attention as predicted class probability and obtain both positive and negative prototypes for pseudo-labeling, and then filter the pseudo-labels via the performance of bag and instance classification.}
\label{fig:combined}
\end{center}
\vskip -0.3in
\end{figure*}

\subsection{Problem Formulation}

\textbf{MIL with Localization.} \,\,
In MIL, a whole slide image (WSI) is represented as a bag $X = [x_1 ; \cdots ; x_M] \in \mathbb{R}^{M \times D}$, where $x_m$ is the $m$-th instance (or patch) within the slide and $M$ varies across slides reaching up to $10^5$. 
$X$ is positive if and only if it contains at least one positive instance; otherwise, it is negative. For training, each bag $X$ is assigned a bag-level label $Y \in \mathcal{Y}_{\text{bag}}$ where $\mathcal{Y}_{\text{bag}}$ is the label space, while instance-level localization labels $y_m \in \{0,1\}$ are unavailable. Then, we train the MIL model to predict both bag-level and instance-level labels. For this, we basically follow attention-based MIL \cite{ilse2018attention, li2021dual, tang2023multiple}, usually decomposed into three functions: an instance feature extractor $\mathcal{F}(\cdot)$, an aggregator into a bag representation $\mathcal{E}(\cdot)$, and a bag classifier $\mathcal{C}(\cdot)$. This pipeline is then expressed as: $( \hat{Y}, \{\hat{y}\}_{m=1}^M ) \leftarrow 
\mathcal{C}\left(\mathcal{E}\left(\mathcal{F}(X)\right)\right)$, where $ \hat{Y}$ and $ \{\hat{y}\}_{m=1}^M $ are the bag-level and instance-level predictions. However, feature extraction for the large number of instances causes serious computational overhead. Thus, several works \cite{shao2021transmil, tang2024feature, bui2024falformer} proposed to first extract instance features from a frozen pre-trained backbone and introduce a re-embedding module $\mathcal{R}(\cdot)$ for task-adaptive refinement:
\begin{align}
    ( \hat{Y}, \{\hat{y}\}_{m=1}^M ) \leftarrow \mathcal{C}\left( \mathcal{E}\left(\mathcal{R}\left(\mathcal{F}(X)\right)\right)\right),
    \label{eq:mil_model}
\end{align}
where $\mathcal{F}$ is frozen and the others are the learnable functions.

\textbf{Continual MIL.} In a realistic scenario, the task that the MIL model learns changes over time, to which it must adapt. We denote
$\mathcal{D} = \{\mathcal{D}_1, \cdots, \mathcal{D}_N\}$ as the sequence of datasets for each task
where the $n$-th dataset $\mathcal{D}_n = \{(X_i^n, Y_i^n)\}$ contains tuples of a bag $X_i^n$ and its bag-level label $Y_i^n$, and $1 \leq n \leq N$ is the task index. When training on current dataset $\mathcal{D}_{n}$, previous datasets (\textit{i.e.}, $\mathcal{D}_1, \cdots, \mathcal{D}_{n-1}$) are unavailable or limited. Then, the continual MIL model in \cref{eq:mil_model} needs to continuously learn new knowledge from the sequence of nonstationary datasets $\mathcal{D}$, while maintaining the previously learned knowledge.


\subsection{Grouped Double Attention Transformer} \label{sec:cdat}

Enhanced localization basically requires refinement of features from $\mathcal{F}$. For feature refinement, recent MIL works have leveraged efficient attentions like linear attention \cite{xiong2021nystromformer, shao2021transmil, bui2024falformer} or multi-level attention \cite{chen2022scaling, tang2024feature}.
However, linear attention struggles to capture diversity among instances \cite{han2025agent} and multi-level attention emphasizes dependencies among instances within a region, leading to limited localization. Instead, we design the Grouped Double Attention Transformer (GDAT) inspired by \cite{han2025agent}, depicted in \cref{fig:combined}.

Let $Q = X W_Q$, $K = X W_K$, and $V = X W_V$ be the query, key, and value with $W_Q, W_K, W_V \in \mathbb{R}^{D \times D_1}$, respectively. The standard attention, $O = \text{Attn}(Q,K,V) \in \mathbb{R}^{M \times D}$, is represented with each feature $O_i \in \mathbb{R}^{D_1}$ as:
\begin{align}
    O_i = \sum_{j=1}^{M} \frac{\sigma(Q_i, K_j)}{\sum_{j=1}^{M} \sigma(Q_i, K_j)} V_j,
    \label{eq:st_self_attn}
\end{align}
where $\sigma(Q,K) = \text{exp} (QK^T/\sqrt{D_2})$. Unfortunately, the quadratic complexity $\mathcal{O}(M^2)$ of \cref{eq:st_self_attn} makes the use of the attention for tens of thousands of instances infeasible.

To reduce the cost while still encoding the global dependency among instances, we reformulate the standard attention into a form applying two computationally efficient attentions, illustrated in \cref{fig:combined}.
For this, we follow \cite{tang2024feature} to group instances into regions resulting in a small number of tokens. Let $X_A \in \mathbb{R}^{m \times D}$ be the average pooled tokens in each region. Given $Q_A = X_A W_Q$ and $K_A = X_A W_K$, we utilize more efficient attention represented as:
\begin{align}
    \tilde{O} = \text{Attn}(Q, K_A, \text{Attn}(Q_A, K, V)),
    \label{eq:double_attn}
\end{align}
where $\tilde{O} \in \mathbb{R}^{M \times D}$. Since the complexity of the two attentions in \cref{eq:double_attn} is $\mathcal{O}(mM)$ each, the overall complexity of \cref{eq:double_attn} is also $\mathcal{O}(mM)$, significantly reducing the costs.

Although \cref{eq:double_attn} resolves the complexity, it represents each instance as a combination of $m$ tokens, which may lead to degraded diversity among instance features. To compensate for this, we add the original values $V$, represented as:
\begin{align}
    \tilde{O} = \text{Attn}(Q, K_A, \text{Attn}(Q_A, K, V)) + \eta V
    \label{eq:cdattn}
\end{align}
where $\eta$ is a hyper-parameter. Our Grouped Double Attention Transformer (GDAT), consists of multiple blocks of \cref{eq:double_attn} and serves as $\mathcal{R}$ in our CoMEL framework.

\subsection{Bag Prototypes-based Pseudo-Labeling} \label{sec:ppl}
Let $A = [a_1, a_2, \cdots, a_M]$ be the attention value for instances in $X$ such that $\sum_{m=1}^M a_{m} = 1$. This attention value is used to merge $x_m$'s into a bag feature and evaluate localization by visualizing the attention or classifying positive/negative instances \cite{li2021dual, chen2022scaling, shao2021transmil, tang2024feature}. 
To improve localization based on $A$, we propose Bag Prototype-based Pseudo Labeling (BPPL), adaptively filtering informative pseudo-labels based on attention quality, depicted in \cref{fig:combined}. 

To classify instances based on the attention value, we first normalize each element in $A$ into $[0,1]$. Min-Max normalization is one of choices, but it exaggerates the differences in attention. For example, even when the bag contains only negative samples,
it enforces at least one instance to have an attention value of either zero or one although it is desirable that all instances have uniform near-zero attention values. Instead, we use $\mathcal{A} = [\alpha_1, \alpha_2, \cdots, \alpha_M]$ as the class probability where $ \alpha_m = (a_m-\min_{i} a_i ) / \max_{i} a_i$. Then, assuming the MIL model can successfully localizes positive instances, their values in $\mathcal{A}$ get close to one while most instances in a negative bag have near-zero attention values. 

To refine the normalized scores $\mathcal{A}$, we obtain a pair of one positive and negative prototype from a bag and use them to pseudo-label the instances. We will apply this only to bags that contain positive instances. 
Specifically, the bag prototype $x_{\text{bag}}$ from the instances in attention-based MIL is determined by $x_{\text{bag}} = \sum_{m=1}^{M} a_{m} x_{m}$. If the MIL model can successfully localize positive instances for a bag, $\alpha_m$ takes high values only for positive instances. Then, we can define a reversed attention $\tilde{A}=[\tilde{a}_1, \tilde{a}_2, \cdots, \tilde{a}_M]$ and the prototype representing negative instances is obtained by:
\begin{align}
    \tilde{x}_{\text{bag}} = \sum_{m=1}^{M} \tilde{a}_{m} x_{m}, \quad \text{s.t.} \,\, \tilde{a}_m = \frac{1-\alpha_m}{\sum_{m=1}^{M} (1-\alpha_m)}.
\end{align}
By measuring similarity of $x_m$ to $x_{\text{bag}}$ and $\tilde{x}_{\text{bag}}$, we can identify positive and negative instances to pseudo-label $x_m$.

Unfortunately, MIL models often struggle with localization in practice. In that case, $x_{\text{bag}}$ and $\tilde{x}_{\text{bag}}$ contain mixed combination of positive and negative instances. To address this, we propose two simple methods to improve the quality of bag prototypes and pseudo-labels. First, we minimize the similarity between $x_{\text{bag}}$ and $\tilde{x}_{\text{bag}}$ during training:
\begin{align}
    \mathcal{L}_{\text{sep}} = \textbf{sim}(x_{\text{bag}}, \tilde{x}_{\text{bag}}),
    \label{eq:loss_sep}
\end{align}
where $ \textbf{sim} $ denotes the cosine similarity.
To further enhance discriminativity, we compare the residual instance features with respect to $\bar{x} = (x_{\text{bag}} +\tilde{x}_{\text{bag}}) / 2 $.
Specifically, we define $r = x_{\text{bag}} - \bar{x}_{\text{bag}}$ and $\tilde{r} = \tilde{x}_{\text{bag}} - \bar{x}_{\text{bag}} $. With these features, we compute the binary class probability for $x_m$ as:
\begin{align}
    p_{m} = \mathbf{S} ( [ \frac{\textbf{sim}(x_m-\bar{x}, r )}{T}, \frac{\textbf{sim}(x_m-\bar{x}, \tilde{r} )}{T} ] ),
\end{align}
where $ \mathbf{S} $ is the softmax and $ T $ is the temperature.  Then, $\hat{y}_m = \arg \max p_m$ is utilized as the pseudo-label. To filter the pseudo-labels, we leverage both the MIL model's bag classification performance and the confidence on instance predictions. With $\hat{p}_m = \max p_m$, we minimize the loss:
\begin{align}
    \mathcal{L}_{m} =  \mathbbm{1} (\hat{Y} = Y) \mathbbm{1} (\hat{p}_m > \tau_1) \mathbb{H} (\hat{y}_m, \alpha_m),
    \label{eq:loss_pseudo}
\end{align}
where $\hat{Y}$ is the predicted bag class, $\mathbbm{1}(\cdot)$ is the indicator function, $\mathbb{H}$ is the cross-entropy, and $\tau_1$ is defined as
$\tau_1 = 0.5 + \tau  \mathbb{E} \left[ \mathbbm{1}(\hat{Y} = Y) \right]$, which is adapted based the performance on bag classification for training dataset. Using \cref{eq:loss_sep} and \cref{eq:loss_pseudo}, we define the instance loss as:
\begin{align}
    \mathcal{L}_{\text{inst}} = \mathbbm{1}(Y_{\text{bag}} \in \mathcal{Y}_{\text{pos}})  \left(\frac{\lambda_1}{M} \sum_{m=1}^{M} \mathcal{L}_{m} + \lambda_2 \mathcal{L}_{\text{sep}}\right),
\end{align}
where $ \mathcal{Y}_{\text{pos}} $ is the label set of positive classes. Note that $ \mathcal{Y}_{\text{pos}} = \{ 1 \} $ for tumor detection while $ \mathcal{Y}_{\text{pos}} = \mathcal{Y}_{\text{bag}} $ for the tumor subtyping task. Thus, the losses $ \mathcal{L}_m $ and $ \mathcal{L}_{\text{sep}} $ are applied only when the bag contains positive instances. Considering both bag and instance classification, we finally optimize $\mathcal{L}_{\text{WSI}} = \mathcal{L}_{\text{bag}} + \mathcal{L}_{\text{inst}}$ for a bag where $\mathcal{L}_{\text{bag}}$ is the cross entropy for bag classification, commonly used by prior arts.


\begin{figure}[t]
\begin{center}
\centerline{\includegraphics[width=0.95\columnwidth]{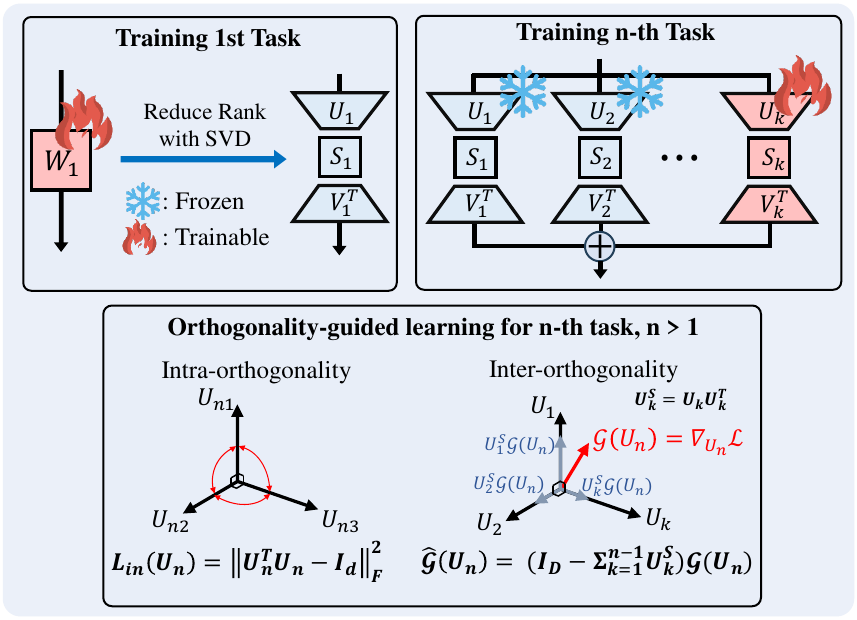}}
 \vskip -0.1in
\caption{Illustration of Orthogonal Weighted Low-Rank Adaptation (OWLoRA) for continual MIL. 
For orthogonality of 1st task, OWLoRA trains the full-rank weights and then extracts principal orthogonal components by singular value decomposition. For subsequent tasks, it imposes the intra- and inter-orthogonality on the bases for low-rank learnable matrices for the current task.
}
\label{fig:owlora}
\vskip -0.4in
\end{center}
\end{figure}

\subsection{Orthogonal Weighted Low-Rank Adaptation} \label{sec:owlora}

ConSlide \cite{huang2023conslide} proposed a rehearsal-based continual MIL method, but it could raise privacy issue, a major concern to medical experts and patients in real hospitals, and focused solely on bag classification.
QPMIL-VL \cite{gou2024queryable} is rehearsal-free, but it also prioritized bag classification and relied on vision-language models \cite{lu2024visual, chen2024towards} trained on large-scale histopathological data.
Instead, we propose a LoRA-based approach \cite{hulora} dubbed Orthogonal Weighted Low-Rank Adaptation (OWLoRA), allowing the use of arbitrary feature extractors and preventing forgetting of both bag and instance classification in rehearsal-free manner, as \cref{fig:owlora}.


Specifically, we train low-rank matrices composed of orthogonal bases to update weights in the MIL model for each task.
Let $W_n \in \mathbb{R}^{D \times D_1}$ represent a linear weight after learning the $n$-th task. Then, we will represent $W_n$ as:
\begin{align} 
    W_n = \sum_{k=1}^{n} U_k S_k (V_k)^T, \nonumber 
\end{align}
where $U_k \in \mathbb{R}^{D \times d}$, $V_k \in \mathbb{R}^{D_1 \times d}$ are matrices whose columns represent orthonormal bases with $d \ll D, D_1$, and $S_k \in \mathbb{R}^{d \times d}$ is a diagonal matrix. Here, $U_k S_k (V_k)^T$ is the matrix trained for the $k$-th task. 
For $n$-th task, we only train $U_n$, $S_n$, and $V_n$, while $U_k$, $S_k$, and $V_k$ for $k < n$ remain frozen. 
Then, we describe how to ensure orthogonality both among the columns within $U_n$ and between $U_n$ and $U_k$'s. Since we will apply the same approach to both $U_k$ and $V_k$, we explain it only for $U_k$ if there's no confusion.

Previous LoRA based CL works \cite{liang2024inflora, qiaolearn} adapted pre-trained weights, which enabled to use LoRA from the first task. However, MIL models require to train the re-embedding module $\mathcal{R}$, aggregator $\mathcal{E}$, and bag classifier $\mathcal{C}$ from the scratch for the first task. Therefore, we train a matrix $W_1 \in \mathbb{R}^{D_1 \times D_2}$ for the first task and then train the weights $U_k S_k (V_k)^T$ for subsequent tasks. To ensure orthogonality between $W_1$ and $U_k S_k (V_k)^T$, we extract only a few principal components with the largest singular values in $W_1$. Let $U_1 S_1 (V_1)^T$ be the the singular value decomposition of $W_1$. We reduce it to $\tilde{U}_1 \tilde{S}_1 (\tilde{V}_1)^T$, where $\tilde{U}_1 \in \mathbb{R}^{D \times \tilde{d}}$ and $\tilde{V}_1 \in \mathbb{R}^{D_1 \times \tilde{d}}$. Then, $\tilde{d}$ is determined by: 
\begin{align} 
    \tilde{d} = \arg \min_{i} \{i \in {1,2, ..., D} \,\, | \,\, \frac{\| S_1[1:i] \|_2^2}{\| S_1 \|_2^2} \ge \epsilon \}, \nonumber 
\end{align} 
where $S_1 = \text{diag}(S_1)$ is the diagonal of $S_1$ and $S_1[1:i]$ is the top-$i$ largest singular values in $S_1$.
For further tasks with $k > 1$, we train $U_k$ to ensure the orthogonality to $\tilde{U}_1$

\begin{table*}[t]
    \centering
    
    \setlength{\tabcolsep}{8pt}
    \renewcommand{\arraystretch}{1.0}
    
    \resizebox{\textwidth}{!}{%
    \begin{tabular}{|c|c|c|c|c|c|c|c|}
        \hline
        \textbf{CL Type} 
        & \textbf{Method} 
        & \textbf{$\text{ACC}_{\text{inst}}$ ($\uparrow$)} 
        & \textbf{$\text{Forget}_{\text{inst}}$ ($\downarrow$)} 
        & \textbf{IoU ($\uparrow$)} 
        & \textbf{Dice} ($\uparrow$) 
        & \textbf{ $\text{ACC}_{\text{bag}}$ ($\uparrow$)} \\
        \hline

        \multirow{2}{*}{Upper Bound}
        
        & Joint (Full label) 
        & $90.94 \pm 2.87$ 
        & -
        & $67.35 \pm 2.54$ 
        & $76.81 \pm 2.62$ 
        & $75.82 \pm 2.31$ \\ 
        
        & Joint (Weak label) 
        & $79.80 \pm 1.87$ 
        & -
        & $49.48 \pm 2.94$  
        & $60.56 \pm 2.47$ 
        & $73.23 \pm 2.84$ \\

        \hline
        
        Lower Bound 
        & Finetune
        & $60.00 \pm 1.13$ 
        & $26.60 \pm 2.28$
        & $10.18 \pm 2.66$ 
        & $15.08 \pm 1.75$ 
        & $13.42 \pm 1.54$ \\
        
        \hline

        \multirow{2}{*}{Regularization-based} 
        & EWC \cite{kirkpatrick2017overcoming}
        & $64.51 \pm 2.53$ 
        & $23.66 \pm 2.57$
        & $11.62 \pm 1.79$ 
        & $14.27 \pm 3.01$ 
        & $16.76 \pm 2.28$ \\
        
        & LwF \cite{li2017learning}
        & $65.86 \pm 2.66$ 
        & $22.81 \pm 2.13$
        & $11.79 \pm 2.37$ 
        & $16.00 \pm 3.03$ 
        & $23.22 \pm 1.56$ \\

        \hline
        
        \multirow{6}{*}{Rehearsal-based} 
        & A-GEM/30 \cite{chaudhry2018efficient} 
        & $66.69 \pm 1.82$ 
        & $20.67 \pm 2.41$
        & $14.50 \pm 3.14$ 
        & $21.95 \pm 1.59$ 
        & $42.85 \pm 2.92$ \\
        
        & ER/30 \cite{rolnick2019experience} 
        & $66.36 \pm 2.63$ 
        & $22.37 \pm 2.13$
        & $13.49 \pm 1.20$ 
        & $20.39 \pm 1.77$ 
        & $47.39 \pm 2.43$ \\

        & ER/100 \cite{rolnick2019experience} 
        & $68.63 \pm 2.32$ 
        & $19.09 \pm 1.74$
        & $20.03 \pm 2.88$ 
        & $27.10 \pm 3.00$ 
        & $50.43 \pm 2.09$ \\

        & DER++/30 \cite{buzzega2020dark}
        & $68.08 \pm 1.93$ 
        & $20.11 \pm 1.88$
        & $19.53 \pm 1.95$ 
        & $26.67 \pm 2.13$ 
        & $49.65 \pm 2.24$ \\

        & ER-ACE/30 \cite{caccianew}
        & $67.41 \pm 2.61$ 
        & $20.51 \pm 2.78$
        & $15.35 \pm 2.86$ 
        & $22.40 \pm 2.66$ 
        & $47.72 \pm 2.13$ \\

        & ConSlide/30 \cite{huang2023conslide} 
        & $66.51 \pm 1.37$ 
        & $19.93 \pm 1.49$
        & $17.23 \pm 2.24$ 
        & $25.73 \pm 2.05$ 
        & $56.73 \pm 1.58$ \\
        
        \hline

        Prompt-tuning-based
        & QPMIL-VL \cite{gou2024queryable} 
        & \underline{$68.75 \pm 1.93$} 
        & $18.64 \pm 1.78$
        & $25.87 \pm 2.05$ 
        & $34.71 \pm 1.90$ 
        & \underline{$59.28 \pm 1.89$} \\
        
        \hline

        \multirow{3}{*}{LoRA-based} 
        & LoRA \cite{hulora} 
        & $64.64 \pm 2.06$ 
        & $21.22 \pm 1.85$
        & $19.85 \pm 2.18$ 
        & $28.03 \pm 2.84$ 
        & $31.59 \pm 1.90$ \\

        & InfLoRA \cite{liang2024inflora}
        & $70.17 \pm 2.45$ 
        & \underline{$17.89 \pm 2.06$}
        & \underline{$31.87 \pm 1.85$}
        & \underline{$40.84 \pm 2.54$}
        & $56.93 \pm 2.18$ \\

        & \textbf{CoMEL (Ours)} 
        & $\textbf{72.64} \pm \textbf{1.78}$ 
        & $\textbf{14.00} \pm \textbf{1.84}$
        & $\textbf{41.87} \pm \textbf{2.44}$ 
        & $\textbf{51.44} \pm \textbf{2.22}$ 
        & $\textbf{62.96} \pm \textbf{1.94}$ \\

        \hline
        
    \end{tabular}%
    }
    
    \caption{Quantitative results of CL methods on instance classification in the continual MIL setup. The best and second best results are marked as \textbf{bold} and \underline{underline}. Each experiment consisted of 10 runs. We conducted the experiments on five sequential organ datasets from combined CM-16  and PAIP. For baselines, we applied the CL approaches upon our GDAT+BPPL, except for ConSlide \cite{huang2023conslide}. All metrics were measured in percentage. CoMEL achieved the highest performance across all metrics while minimizing the forgetting.}

    \label{tab:quant_continual_instance}

    \vskip -0.1in
    
\end{table*}

For training $n$-th task while ensuring orthogonality in $\{U_k\}_{k=1}^n$, we employ two strategies: intra-orthogonality among columns in $U_n$ and inter-orthogonality between $U_n$ and $U_k$ for $k < n$.
For intra-orthogonality, we minimize:
\begin{align}
    \mathcal{L}_{\text{in}} (U_n, V_n) = \| (U_n)^T U_n - I_d\|_F^2 + \| (V_n)^T V_n - I_d\|_F^2, \nonumber
\end{align}
where $\| \cdot \|_F$ is the Frobenius norm and $I_d$ is the identity matrix. That is, we ensure that each column of $U_n$ has a norm of one and the inner product between columns is zero, forming an orthonormal basis. Then, loss for training the $n$-th task considering the intra-orthogonality is defined as:
\begin{align} 
    \mathcal{L} = \mathcal{L}_{\text{WSI}}(U_n, S_n, V_n) + \lambda_3 \mathcal{L}_{\text{in}} (U_n, V_n) 
    \label{eq:final_loss} 
\end{align}

For inter-orthogonality, we project the gradients of $U_n$ obtained from $\mathcal{L}$ onto the space orthogonal to the spans of $U_k$'s for $k < n$. Let $\mathcal{G}(U_n) = \nabla_{U_n} \mathcal{L}$ be the gradient of $\mathcal{L}$ for $U_n$. Then, we project $\mathcal{G}(U_n)$ as: 
\begin{align}
    \hat{\mathcal{G}}(U_n) =  \left( I_D - \sum_{k=1}^{n-1} U_k (U_k)^T \right) \mathcal{G}(U_n).
    \label{eq:inter_projection} 
\end{align}
Then, we update $U_n$ by the gradient descent $U_n \leftarrow U_n - \gamma \hat{\mathcal{G}}(U_n)$. 
Thus, by leveraging \cref{eq:final_loss} and \cref{eq:inter_projection}, we enable the model to learn new tasks both enhancing instance classification and preventing forgetting of previous tasks.

\begin{figure}[t]
\begin{center}
\centerline{\includegraphics[width=1.0\columnwidth]{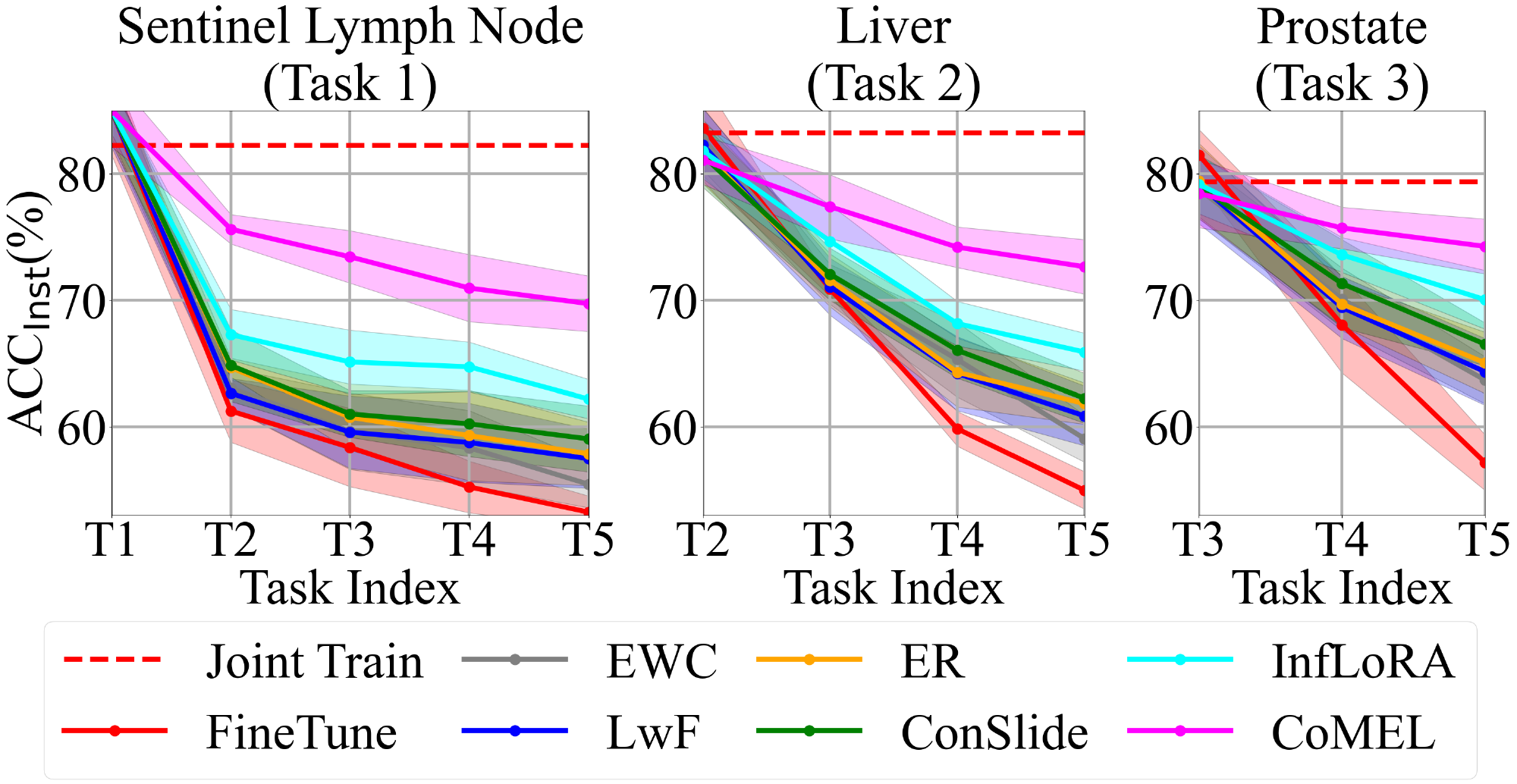}}
 \vskip -0.1in
\caption{Instance-level classification accuracy (ACC$_\text{inst}$) over sequentially learned organs from combined CM-16 and PAIP. Each figure shows the forgetting for a specific organ under continual MIL setup. Task Index indicates the order of organs in the sequential tasks. CoMEL consistently achieved superior performance across all tasks, effectively mitigating catastrophic forgetting.}
\label{fig:graph_continual_instance}
\vskip -0.3in
\end{center}
\end{figure}

\section{Experiements}
\subsection{Experimental Setups}

\begin{figure*}[t]
\begin{center}
\centerline{\includegraphics[width=1.0\textwidth]{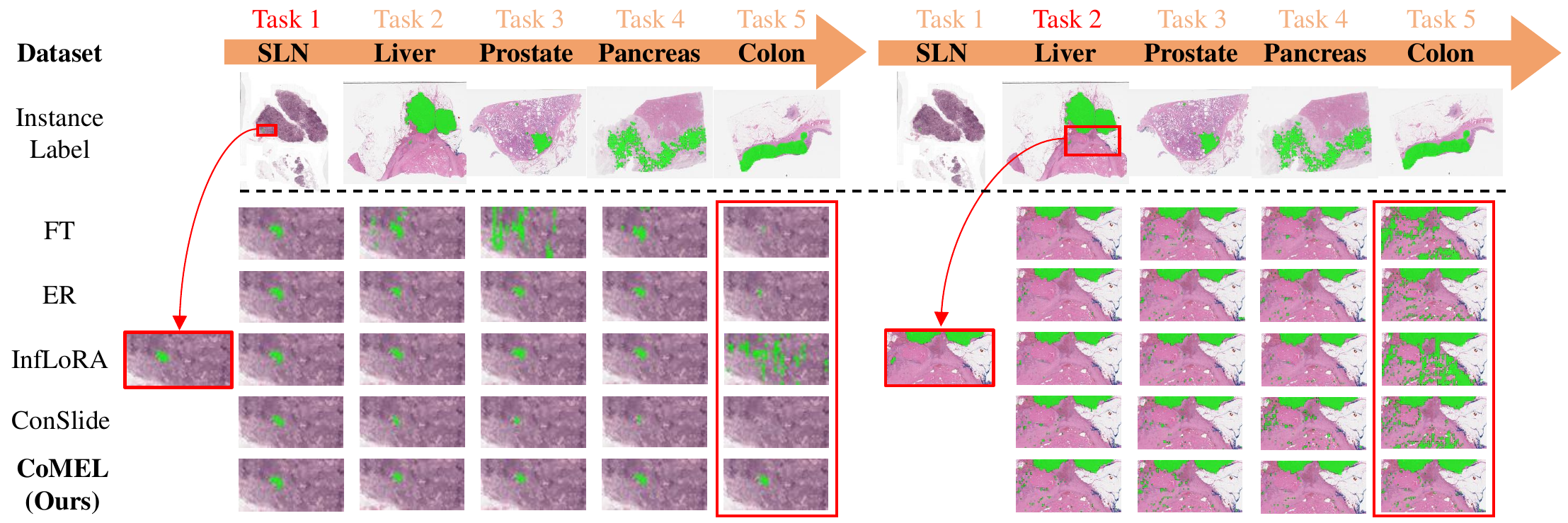}}
 \vskip -0.05in
\caption{Qualitative results of localization across sequential organ datasets under continual MIL setup. Each column shows localization performance on Task 1 (left) or 2 (right) as the learned organ changes over sequential tasks. Each row corresponds to CL methods including CoMEL. CoMEL successfully preserved the localization quality across all tasks, while baselines increase false positives or false negatives.}
\label{fig:qualitative_continaul_instance}
\vskip -0.4in
\end{center}
\end{figure*}


\begin{table}[t]
    \centering

    \setlength{\tabcolsep}{3pt}
    \renewcommand{\arraystretch}{1.1}
    
    \resizebox{\columnwidth}{!}{%
    \begin{tabular}{|c|c|c|c|}
        \hline
        \textbf{Method} & \textbf{$\text{ACC}_{\text{bag}}$ ($\uparrow$)} & \textbf{$\text{Forget}_{\text{bag}}$ ($\downarrow$)} &  \textbf{M.$\text{ACC}_{\text{bag}}$ ($\uparrow$)} \\
        \hline

        Joint
        & $91.49 \pm 2.28$ & -  & $93.31 \pm 2.67$ \\

        \hline
        
        Finetune
        & $32.40 \pm 3.49$ & $64.50 \pm 2.34$ & $83.67 \pm 3.40$ \\
        
        \hline
        
        EWC \cite{kirkpatrick2017overcoming}
        & $36.77 \pm 4.07$ & $62.72 \pm 2.36$ & $83.91 \pm 3.29$ \\
        
        LwF \cite{li2017learning}
        & $37.91 \pm 4.41$ & $61.92 \pm 4.97$ & $88.65 \pm 2.14$ \\
        
        \hline
        
        A-GEM/30 \cite{chaudhry2018efficient} 
        & $47.24 \pm 4.64$ & $57.43 \pm 4.50$ & $89.27 \pm 2.44$ \\
        
        ER/30 \cite{rolnick2019experience} 
        & $80.57 \pm 4.83$ & $17.19 \pm 3.02$ & $89.62 \pm 3.21$ \\

        ER/100 \cite{rolnick2019experience} 
        & $84.73 \pm 3.95$ & $14.64 \pm 3.83$ & $90.29 \pm 3.96$ \\
        
        DER++/30 \cite{buzzega2020dark}
        & $81.01 \pm 2.37$ & $15.26 \pm 3.15$ & $89.69 \pm 4.67$ \\
        
        ConSlide/30 \cite{huang2023conslide} 
        & $84.88 \pm 4.98$ & $14.35 \pm 3.22$ & $90.73 \pm 3.12$ \\

        \hline

        QPMIL-VL \cite{gou2024queryable} 
        & $\underline{86.53 \pm 3.72}$ & $\underline{12.17 \pm 3.60}$ & $\underline{91.07 \pm 3.12}$ \\

        \hline

        LoRA \cite{hulora} 
        & $50.16 \pm 3.93$ & $32.66 \pm 4.58$ & $87.79 \pm 2.58$ \\
        
        InfLoRA \cite{liang2024inflora}
        & $82.47 \pm 2.85$ & $15.71 \pm 3.65$ & $89.65 \pm 4.04$ \\
        
        \textbf{CoMEL (Ours)} 
        & $\mathbf{87.94 \pm 3.69}$ & $\mathbf{10.12 \pm 2.77}$ & $\mathbf{92.28 \pm 3  .10}$ \\
        
        \hline
    \end{tabular}%
    }
    \vskip -0.05in
    \caption{Comparison of CL methods for continual bag classification on TCGA dataset. The best and second best results are marked as \textbf{bold} and \underline{underline}. Each experiment consisted of 3 runs.}
    \label{tab:quant_continual_bag}
    \vskip -0.2in
\end{table}


\textbf{Datasets.} \,\, For continual MIL experiments, we conducted two types of experiments: 1) continual instance classification for tumor detection and 2) continual bag classification for tumor subtyping.
For the continual instance classification, we constructed a sequence of multi-organ datasets with dense instance-level annotations. Specifically, we combined \textbf{CAMELYON-16 (CM-16)} \cite{bejnordi2017diagnostic} and \textbf{Pathology AI Platform (PAIP)} \cite{kim2021paip, kim2023paip, wisepaip}. CM-16 contains the tumor segmentation dataset for sentinel lymph nodes. PAIP provides tumor segmentation dataset for multi-organs. From PAIP, we used the datasets for liver, prostate, pancreas, and colon. The datasets are summarized in \cref{tab:cm16_paip_statistics}. Since the PAIP dataset only includes tumor slides, we randomly split the slides for each organ into two subsets as tumor and normal slides. For the normal slides, we constructed bags by extracting only normal patches without any tumor pixels. Then, we sequentially trained on the five organ datasets in the following order: sentinel lymph nodes (SLN), liver, prostate, pancreas, and colon.
For continual bag classification, we followed the setup of ConSlide \cite{huang2023conslide} and used four datasets from \textbf{The Cancer Genome Atlas (TCGA)} \cite{weinstein2013cancer}: non-small cell lung carcinoma (NSCLC), invasive breast carcinoma (BRCA), renal cell carcinoma (RCC), and esophageal carcinoma (ESCA), shown in \cref{tab:tcga_data_statistics}. The task sequence is: NSCLC, BRCA, RCC, and ESCA.

\textbf{Evaluation metrics.} \,\, 
We evaluated the model's final averaged performance on the learned datasets using instance-level accuracy (ACC$_{\text{inst}}$), Intersection over Union (IoU), Dice score, and bag-level accuracy (ACC$_{\text{bag}}$). To evaluate the forgetting of localization, we used the forgetting measure (Forget$_{\text{inst}}$) based on ACC$_{\text{inst}}$. We also measured the model's final averaged performance using bag-level accuracy (ACC$_{\text{bag}}$), the bag-level forgetting measure (Forget$_{\text{bag}}$), and computed the masked bag-level accuracy (M.ACC\textsubscript{bag}). Here, M.ACC\textsubscript{bag} is evaluated only considering classes belonging to the same task, following ConSlide \cite{huang2023conslide}. Details on these metrics can be found in Sec. \ref{sec:further_metrics}.

\begin{table*}[t]
    \centering
    \setlength{\tabcolsep}{8pt}
    \renewcommand{\arraystretch}{1.0}
    
    \resizebox{\textwidth}{!}{%
    \begin{tabular}{|c|c|c|c|c|c|c|}
        \hline
        
        \textbf{Model} 
        & \textbf{$\text{ACC}_{\text{bag}}$ ($\uparrow$)} 
        & \textbf{$\text{AUC}_{\text{bag}}$ ($\uparrow$)} 
        & \textbf{$\text{F1}_{\text{bag}}$ ($\uparrow$)} 
        & \textbf{$\text{ACC}_{\text{inst}}$ ($\uparrow$)} 
        & \textbf{IoU ($\uparrow$)} 
        & \textbf{Dice ($\uparrow$)} \\
        
        \hline
        
        ABMIL \cite{lu2021data} 
        & $68.67 \pm 0.75$ & $93.61 \pm 0.85$
        & $68.13 \pm 1.54$ & $65.87 \pm 1.24$ 
        & $33.55 \pm 1.73$ & $47.23 \pm 0.74$ \\
        
        DS-MIL \cite{li2021dual}
        & $68.66 \pm 0.50$ & $93.08 \pm 0.68$
        & $68.71 \pm 1.48$ & $63.44 \pm 0.90$ 
        & $30.65 \pm 0.30$ & $41.14 \pm 0.54$ \\
        
        TransMIL \cite{shao2021transmil} 
        & $69.29 \pm 0.23$ & $93.42 \pm 0.75$
        & $68.67 \pm 0.80$ & $58.30 \pm 1.27$ 
        & $25.76 \pm 1.59$ & $37.06 \pm 1.48$ \\

        RRT-MIL \cite{tang2024feature} 
        & $70.17 \pm 1.04$ & $91.44 \pm 0.69$
        & $69.12 \pm 1.58$ & $55.28 \pm 0.80$ 
        & $26.80 \pm 0.22$ & $37.95 \pm 1.30$ \\

        smAP \cite{castrosm} 
        & $68.67 \pm 1.28$ & $94.56 \pm 0.52$
        & $68.35 \pm 0.27$ & $68.24 \pm 1.23$ 
        & $36.45 \pm 1.48$ & $49.33 \pm 1.79$ \\

        \hline

        \textbf{GDAT (Ours)} 
        & $70.80 \pm 0.61$ & $93.41 \pm 0.39$
        & $71.31 \pm 0.64$ & $68.36 \pm 0.58$ 
        & $36.31 \pm 0.62$ & $48.55 \pm 1.15$ \\

        \textbf{GDAT+BPPL (Ours)} 
        & $71.87 \pm 0.83$ & $94.20 \pm 1.13$
        & $71.78 \pm 0.99$ & $79.27 \pm 0.61$ 
        & $49.32 \pm 0.96$ & $60.68 \pm 1.06$ \\
        
        \hline
    \end{tabular}%
    }
    
    \vskip -0.05in    
    \caption{Comparison of different MIL models on the merged dataset, combining five organs from CM16 and PAIP. For bag-level evaluation, we measured ACC$_{\text{bag}}$, AUC$_{\text{bag}}$, and F1$_{\text{bag}}$. For instance-level evaluation, we used ACC$_{\text{inst}}$, IoU, and Dice score. The best and the second best results are marked as \textbf{bold} and \underline{underline}. Each experiment consisted of 10 runs. The proposed method achieved the best performance across all metrics, demonstrating its superiority in both bag-level and instance-level classification.}
    \label{tab:quant_joint_instance}
    \vskip -0.1in
\end{table*}

\begin{table}[t]
    \centering

    \setlength{\tabcolsep}{5pt}
    \renewcommand{\arraystretch}{1.0}
    
    \resizebox{\columnwidth}{!}{%
    \begin{tabular}{|c|c|c|c|}
        \hline
        \textbf{Model}
        & \textbf{$\text{ACC}_{\text{inst}}$ ($\uparrow$)} 
        & \textbf{$\text{Forget}_{\text{inst}}$ ($\downarrow$)} 
        & \textbf{Dice ($\uparrow$)} \\
        \hline
        
        ResNet50 \cite{he2016deep} 
        & $65.63 \pm 1.81$ & $16.23 \pm 2.04$ & $46.06 \pm 1.60$ \\
        
        PLIP \cite{huang2023visual}
        & \underline{$67.14 \pm 2.31$} & $16.60 \pm 1.58$ & \underline{$48.49 \pm 2.37$} \\
    
        CONCH \cite{lu2024visual}
        & $62.33 \pm 1.58$ & \underline{$14.63 \pm 2.03$} & $46.26 \pm 1.65$ \\
    
        UNI \cite{chen2024towards} 
        & $\textbf{72.64} \pm \textbf{1.78}$ 
        & $\textbf{14.00} \pm \textbf{1.84}$
        & $\textbf{51.44} \pm \textbf{2.22}$ \\
        
        \hline
    \end{tabular}
    }

    \vskip -0.05in

    \caption{Continual instance classification with different feature extractors with 10 runs. 
    While CoMEL performs the best with UNI \cite{chen2024towards}, it achieved robust results with other backbones.}
        
    \label{tab:quant_extractors}

    \vskip -0.1in
    
\end{table}

\textbf{Implementation details.} \,\, We extracted patches of size $256 \times 256$ at 20x magnification for slides from PAIP and TCGA. For CM-16, we first extracted $512 \times 512$ patches at 40x magnification and downsampled them to $256 \times 256$. As the feature extractor $\mathcal{F}$, we utilized UNI \cite{chen2024towards}, a foundation model for computational pathology. For all tasks, we applied the same values for hyper-parameters in CoMEL. We set $\eta=0.1$ in GDAT.
For BPPL, we set $(T,\tau,\lambda_1, \lambda_2)=(0.5,0.35,0.5,1.0)$. For OWLoRA, we set $(\epsilon, d, \lambda_3)=(0.99,16,1.0)$. Further details can be found in Sec. \ref{sec:further_implementation}.

\subsection{Results on Continual Instance Classification}
We evaluated CoMEL against following CL baselines: EWC \cite{serra2018overcoming}, LwF \cite{li2017learning}, A-GEM \cite{chaudhry2018efficient}, ER \cite{rolnick2019experience}, DER++ \cite{bui2024falformer}, ConSlide \cite{huang2023conslide}, LoRA \cite{hulora}, and InfLoRA \cite{liang2024inflora}. We also evaluated training on merged CM-16 and PAIP dataset (Joint) as the upper bound and sequentially training without any CL techniques (Finetune) as the lower bound. For experiment, GDAT+BPPL were applied to all baselines except for ConSlide, since ConSlide has its own MIL framework.

Table \ref{tab:quant_continual_instance} presents the quantitative results for tumor detection across the sequential datasets of combined CM16 and PAIP. From the results for ACC$_{\text{bag}}$ and ACC$_{\text{inst}}$, we observed that while rehearsal-based approaches mitigated forgetting in bag classification with better performance than regularization-based approaches, they still suffered from substantial forgetting in instance classification.
On the other hand, InfLoRA outperformed them in both ACC$_{\text{inst}}$ and Forget$_{\text{inst}}$ with competitive performance in terms of ACC$_{\text{bag}}$. Based on LoRA-based CL approach, CoMEL achieved the best performance across all metrics. In particular, CoMEL outperformed in terms of IoU and Dice by a large margin, demonstrating its effectiveness in preserving instance-level localization under the continual MIL setup.

Figure. \ref{fig:graph_continual_instance} illustrates the instance accuracy (ACC$_\text{inst}$) over sequential tasks. All baselines exhibited severe forgetting as new tasks (organs) were introduced. While regularization- and rehearsal-based methods slightly mitigated the forgetting of localization, InfLoRA could mitigate the forgetting of localization better compared to other baselines. Meanwhile, CoMEL showed the best results in mitigating the forgetting of localization, consistently maintaining the highest accuracy. 
Figure \ref{fig:qualitative_continaul_instance} presents qualitative localization results (green areas) under continual MIL setup. Each column represents the localization performance on SLN (Task 1) or liver (Task 2) over the sequentially learned tasks. 
Baselines suffered from either localization leakage (false positives) or localization disappearance (false negatives). In contrast, CoMEL successfully preserved the localized tumor regions after training on all five datasets.
For further results on continual instance classification, we refer to Sec. \ref{sec:inst_reversed} and \ref{sec:quali_additional}.

\subsection{Results on Continual Bag Classification}
To evaluate the performance on continual bag classification, we compared CoMEL against the same baselines in the continual instance classification. \cref{tab:quant_continual_bag} illustrates the performance of CL baselines and CoMEL for continual tumor subtyping using TCGA dataset. CoMEL achieved the highest performance in terms of ACC\textsubscript{bag} and Forget\textsubscript{bag}, demonstrating its strong performance while effectively mitigating forgetting. Consistently, CoMEL also achieved the highest M.ACC\textsubscript{bag} compared to the baselines.
Rehearsal-based approaches such as ER and ConSlide demonstrated their effectiveness in mitigating catastrophic forgetting for continual bag classification, while regularization-based methods suffered from severe forgetting. For further results on continual bag classification, we refer to Sec. \ref{sec:bag_reversed}.

\subsection{Results on Joint Bag and Instance Classification}
We also conducted experiments for joint bag and instance classification to compare with other MIL models. To this end, we merged all datasets introduced in the continual instance classification, training them all at once. For bag-level evaluation, we measured ACC\textsubscript{bag}, Area Under Curve (AUC\textsubscript{bag}), and F1 Score (F1\textsubscript{bag}). For instance-level evaluation, we measured ACC\textsubscript{inst}, IoU, and Dice score.
We compared our GDAT+BPPL with ABMIL \cite{ilse2018attention}, DS-MIL \cite{li2021dual}, TransMIL \cite{shao2021transmil}, RRT-MIL \cite{tang2024feature}, and smAP \cite{castrosm}. From \cref{tab:quant_joint_instance}, the proposed method achieved the best performance across all metrics. While the advanced MIL models improved bag classification, they provided little improvement on the localization. smAP improved the localization, but GDAT+BPPL proved to be more beneficial. For results on bag and instance classification on single dataset, we refer to Sec. \ref{sec:inst_reversed}.


\begin{table}[t]
    \centering

    \setlength{\tabcolsep}{7pt}
    \renewcommand{\arraystretch}{1.05}
    
    \resizebox{\columnwidth}{!}{%
    \begin{tabular}{|c|c|c|c|c|}
        \hline

        \textbf{$\mathcal{L}_{\text{in}}$ } &
        \textbf{$\hat{\mathcal{G}}(V_n; E)$} &
        \textbf{$\text{ACC}_{\text{inst}}$} ($\uparrow$) & 
        \textbf{$\text{Forget}_{\text{inst}}$} ($\downarrow$) &
        \textbf{$\text{ACC}_{\text{bag}}$} ($\uparrow$) \\

        \hline 
        
        \checkmark & \checkmark
        & \textbf{74.15} $\pm$ \textbf{1.97} & \textbf{13.05} $\pm$ \textbf{2.31} & \textbf{62.64} $\pm$ \textbf{2.14} \\

        $\times$ & \checkmark
        & $66.98 \pm 3.19$ & $20.10 \pm 2.86$ & $38.23 \pm 2.05$ \\
        
        \checkmark & $\times$
        & $71.02 \pm 1.63$ & $17.23 \pm 1.90$ & $60.34 \pm 1.90$ \\
        
        \hline
    \end{tabular}%
    }
    \vskip -0.05in
    
    \caption{Ablation studies of OWLoRA for continual instance classification, showing all components are effective for adaptability (3 runs). The row with \textbf{bold} is the selected configurations.}

    \label{tab:ablation_owlora}

    \vskip -0.1in
    
\end{table}


\subsection{Results with Various Backbones}
To evaluate the robustness of our CoMEL across different feature extractors, we measured its performance using other pre-trained backbones for feature extractors.
Table. \ref{tab:quant_extractors} shows the results of continual instance classification using other feature extractors such as Resnet50 \cite{he2016deep}, PLIP \cite{huang2023visual}, and CONCH \cite{lu2024visual}. 
It demonstrates that CoMEL achieved robust results with the other backbones. We also note that regardless of the types of feature extractors, CoMEL outperformed other CL baselines in Table \ref{tab:quant_continual_instance} in terms of Dice score.  


\subsection{Ablation Studies}
From Table \ref{tab:quant_joint_instance}, the removal of BPPL from GDAT resulted in a notable performance drop especially on the instance classification. Meanwhile, ablating GDAT, leading to the results of ABMIL, verifies that it contributed to improve the performance on bag classification.
We also investigated the impact of each component in OWLoRA like the loss $\mathcal{L}_{\text{in}}$ and the gradient projection $\hat{G}(V_n; E)$. Table~\ref{tab:ablation_owlora} shows that removing $\mathcal{L}_{\text{in}}$ or $\hat{G}(V_n; E)$ significantly degraded all metrics, indicating that both components are essential for continual instance classification. We refer to Sec. \ref{further_ablations} for more discussions and additional ablation studies for CoMEL.

\section{Conclusion}
In this work, we proposed CoMEL, a framework of continual multiple instance learning (MIL) with enhanced localization to address both instance and bag classification, especially for whole slide image analysis. For scalability and localization, we proposed Grouped Double Attention Transformer (GDAT) and Bag Prototypes based Pseudo-Labeling (BPPL). For adaptability to new tasks in continual MIL, we designed Orthogonal Weighted Low-Rank Adaptation (OWLoRA), a rehearsal-free approach imposing orthogonality between learned bases.
Through extensive experiments on public WSI datasets, CoMEL outperformed prior arts in both bag and instance classification while minimizing forgetting under continual MIL setup.

\section*{Acknowledgments}

This work was supported in part by Institute of Information \& communications Technology Planning \& Evaluation (IITP) grant funded by the Korea government(MSIT) [NO.RS-2021-II211343, Artificial Intelligence Graduate School Program (Seoul National University)], a grant of the Korea Health Technology R\&D Project through the Korea Health Industry Development Institute (KHIDI), funded by the Ministry of Health \& Welfare, Republic of Korea (grant number: HI18C0316), the National Research Foundation of Korea(NRF) grant funded by the Korea government(MSIT) (No. RS-2022-NR067592) and AI-Bio Research Grant through Seoul National University. Also, the authors acknowledged the financial support from the BK21 FOUR program of the Education and Research Program for Future ICT Pioneers, Seoul National University.

{
    \small
    \bibliographystyle{ieeenat_fullname}
    \bibliography{comel_main}
}



\clearpage


\twocolumn[{
    \begin{center}
        {\Large \textbf{Continual Multiple Instance Learning with Enhanced Localization \\ for Histopathological Whole Slide Image Analysis} \\ \emph{\textbf{Supplementary Material}}}
    \end{center}
    \vspace{1cm}
}]

\renewcommand{\thesection}{S\arabic{section}}
\renewcommand{\thefigure}{S\arabic{figure}}
\renewcommand{\thetable}{S\arabic{table}}
\renewcommand{\theequation}{S\arabic{equation}}

\setcounter{section}{0}
\setcounter{figure}{0}
\setcounter{table}{0}
\setcounter{equation}{0}

\section{Experimental Setups}

\subsection{Details on Datasets}

We conducted tumor detection using the CAMELYON-16 (CM-16)~\cite{bejnordi2017diagnostic} and Pathology AI Platform (PAIP)~\cite{kim2021paip, kim2023paip, wisepaip} datasets to evaluate continual and joint instance classification. For continual bag classification, we conducted tumor subtyping using The Cancer Genome Atlas (TCGA)~\cite{weinstein2013cancer}.

\textbf{CAMELYON-16 (CM-16)} is a WSI dataset for diagnosing breast cancer metastases in sentinel lymph nodes. It consists of 400 WSIs with corresponding pixel-level tumor annotations, officially split into 270 training slides and 130 test slides. Following~\cite{chen2022scaling, lu2021data}, we merged the official training and test sets and performed three-times threefold cross-validation to ensure that each slide is used for both training and testing. This cross-validation strategy helps mitigate the impact of data partitioning and random seed selection on model evaluation. The numbers of tumor and non-tumor slides in CM-16 are summarized in \cref{tab:cm16_paip_statistics}.


\begin{table}[h]
    \centering

    \setlength{\tabcolsep}{6pt}
    \renewcommand{\arraystretch}{1.0}
    
    \resizebox{\columnwidth}{!}{%
    \begin{tabular}{|c|c|c|c|}
        \hline
        \textbf{Repository} & \textbf{Organ} & \textbf{\# normal slides} & \textbf{\# tumor slides} \\  
        \hline
        
        CM-16 & Sentinel Lymph Nodes & 241 & 159  \\
        \hline
        
        \multirow{4}{*}{PAIP} 
        & Liver & 251 & 252 \\
        
        & Prostate & 299 & 300 \\
        
        & Pancreas & 207 & 207 \\
        
        & Colon & 449 & 449 \\
        
        \hline
    \end{tabular}
    }

    \caption{Datasets for continual instance classification for WSI tumor detection, constructed by organ datasets from CAMELYON-16 (CM-16) \cite{bejnordi2017diagnostic} and Pathology AI Platform (PAIP) \cite{kim2021paip, kim2023paip, wisepaip}.}

    \label{tab:cm16_paip_statistics}

    \vskip -0.1in
    
\end{table}


\textbf{Pathology Artificial Intelligence Platform (PAIP)} is a platform for developing learning-based models for WSI analysis, particularly for tumor diagnosis.  
PAIP consists of hundreds of WSIs across six different organs, each with corresponding pixel-level tumor annotations. Among the available organ datasets, we utilized the liver, prostate, pancreas, and colon datasets. However, these datasets only provide tumor slides, meaning that all slide-level labels correspond to the tumor class. Since application to a MIL setup requires to leverage both tumor and normal slide-level annotations as weak labels, we exploited the MIL formulation where each slide is treated as a bag of multiple instances (patches). Specifically, for each organ dataset in PAIP, we randomly split the slides into two halves — one half designated as tumor slides, and the other as normal slides. For the normal slides, we removed all tumor regions prior to patch extraction to ensure they only contain normal patches. The resulting numbers of tumor and normal slides for each organ in PAIP are summarized in \cref{tab:cm16_paip_statistics}. Similar to CAMELYON-16, we applied three-times threefold cross-validation to each organ dataset to mitigate the effect of data partitioning and random seed selection on model evaluation.

\textbf{The Cancer Genome Atlas (TCGA)} is a large-scale research project jointly conducted by the National Cancer Institute (NCI) and the National Human Genome Research Institute (NHGRI). It aims to systematically analyze genomic alterations in various types of cancer. TCGA provides WSIs from diverse organs along with information on corresponding tumor subtypes, enabling weakly-supervised tumor subtyping tasks. For continual bag classification, we conducted continual tumor subtyping tasks across four organs from TCGA like NSCLC, BRCA, RCC, and ESCA, following the setup of ConSlide~\cite{huang2023conslide}. Detailed statistics for each dataset are summarized in \cref{tab:tcga_data_statistics}.

\begin{table}[h]
    \centering

    \setlength{\tabcolsep}{6pt}
    \renewcommand{\arraystretch}{1.0}
    \resizebox{\columnwidth}{!}{%
    \begin{tabular}{|c|c|c|}
        \hline
        
        \textbf{Dataset} & \textbf{Tumor type} & \textbf{\# slides} \\
        \hline
        \multirow{2}{*}{NSCLC} & Lung adenocarcinoma (LUAD) & 492 \\
                               & Lung squamous cell carcinoma (LUSC) & 466 \\
        \hline
        \multirow{2}{*}{BRCA}  & Invasive ductal (IDC) & 726 \\
                               & Invasive lobular carcinoma (ILC) & 149 \\
        \hline
        \multirow{2}{*}{RCC}   & Clear cell renal cell carcinoma (CCRCC) & 498 \\
                               & Papillary renal cell carcinoma (PRCC) & 289 \\
        \hline
        \multirow{2}{*}{ESCA}  & Esophageal adenocarcinoma (ESAD) & 65 \\
                               & Esophageal squamous cell carcinoma (ESCC) & 89 \\
        \hline
        
    \end{tabular}
    }
    
    \caption{Datasets for continual bag classification for WSI tumor subtyping tasks, constructed by organ datasets from The Cancer Genome Atlas (TCGA) \cite{weinstein2013cancer}.}

    \label{tab:tcga_data_statistics}
\end{table}

\subsection{Details on Evaluation Metrics} \label{sec:further_metrics}

\subsubsection{Continual Instance Classification.}

\textbf{Instance-level accuracy (Acc\textsubscript{inst})} calculates the average instance-level accuracy across tasks after completing the training of all continual tasks. Let $R^{\text{inst}}_{nl}$ denote the instance-level accuracy on the $l$-th task after training on the $n$-th task. For a total of $N$ continual tasks, Acc\textsubscript{inst} is computed as:
\begin{align}
    \text{Acc}_{\text{inst}} = \frac{1}{N}\sum_{l=1}^{N} R^{\text{inst}}_{Nl} 
\end{align}

\textbf{Intersection over Union (IoU)} and \textbf{Dice score} after the final task in continual MIL can be measured in a similar manner.  
Let $\text{IoU}_{nl}$ and $\text{Dice}_{nl}$ denote the IoU and Dice score, respectively, on the $l$-th task after training on the $n$-th task. Then, the IoU and Dice scores after completing all $N$ continual tasks are measured as:

\begin{align}
    \text{IoU} = \frac{1}{N}\sum_{l=1}^{N} \text{IoU}_{Nl}, \quad \text{Dice} = \frac{1}{N}\sum_{l=1}^{N} \text{Dice}_{Nl}
\end{align}

\textbf{Forget on instance-level accuracy (Forget\textsubscript{inst})} quantifies the degree of forgetting of the MIL model on previously learned knowledge as new tasks are introduced, measured in terms of instance-level accuracy. For its task, it measures the gap between the best performance of the task attained during training on sequential tasks and the final performance on the same task after training on all subsequent tasks. Then, it averages the measures over all tasks as:
\begin{align}
\text{Forget}_{\text{inst}} = \frac{1}{N - 1} \sum_{l=1}^{N-1} \max_{n \in \{l, \cdots, N-1\}} R^{\text{inst}}_{n,l} - R^{\text{inst}}_{N,l}.
\end{align}

\subsubsection{Continual Bag Classification.}
\textbf{Bag-level accuracy (Acc\textsubscript{bag})}, similar to continual instance classification, measures the bag-level accuracy after training on all $N$ continual tasks, which is defined as:
\begin{align}
    \text{Acc}_{\text{bag}} = \frac{1}{N}\sum_{l=1}^{N} R^{\text{bag}}_{Nl} 
\end{align}
where $R^{\text{bag}}_{nl}$ is the bag-level accuracy on the $l$-th task after training on the $n$-th task.

\textbf{Forget on bag-level accuracy (Forget\textsubscript{bag})} is defined analogously to Forget\textsubscript{inst} as:
\begin{align}
\text{Forget}_{\text{inst}} = \frac{1}{N - 1} \sum_{l=1}^{N-1} \max_{n \in \{l, \cdots, N-1\}} R^{\text{bag}}_{n,l} - R^{\text{bag}}_{N,l}.
\end{align}

\textbf{Masked bag-level accuracy (M.Acc\textsubscript{bag})} measures the average accuracy computed by restricting classification to only the classes within a task when the task index is given at test time.  
Let $R^{l, \text{bag}}_{Nl}$ be the bag-level accuracy of $l$-th task after training on $N$-th task measured within it class set, under the assumption that the test class is known to belong to the $l$-th task. Then, M.Acc\textsubscript{bag} can be represented as:
\begin{align}
    \text{M.Acc}_{\text{bag}} = \frac{1}{N}\sum_{l=1}^{N} R^{l, \text{bag}}_{Nl} 
\end{align}

\subsection{Further Implementation Details} \label{sec:further_implementation}
We followed CLAM \cite{lu2021data} for patch and feature extraction. Instead of utilizing ResNet-50 like previous works \cite{ilse2018attention, li2021dual, zhang2022dtfd, tang2023multiple}, we utilized UNI \cite{chen2024towards}, the foundation model for computational pathology, as pre-trained feature extractor since it provides improved patch-wise representation for a WSI, resulting in overall enhanced localization results across all MIL and CL methods.

For training continual MIL, we trained each task for 100 epochs.  
For optimization, we adopted Adam~\cite{kingma2014adam} with an initial learning rate of $1 \times 10^{-4}$ and a weight decay of $1 \times 10^{-5}$. To adjust the learning rate during training, we applied a cosine annealing schedule.  
Due to the large size of whole slide images (WSIs), we used a batch size of 1.  
All experiments for both continual and MIL approaches were conducted on a single NVIDIA A6000 GPU.

\section{Further Discussions on CoMEL} \label{further_ablations}

\subsection{Efficiency studies}

\cref{tab:scalability_resource_gdat} compares the computation and memory efficiency of GDAT, RRT-MIL, and smAP. Compared to RRT-MIL, GDAT achieves improved VRAM memory consumption, FLOPs, and inference latency, all with a comparable number of learnable parameters. Note that from \cref{tab:quant_continual_instance}, GDAT outperforms RRT-MIL not only in terms of MIL with localization performance but also in terms of memory and computational efficiency. While smAP exhibits lower memory consumption than GDAT, it suffers from significantly higher inference latency due to the computation of the adjacency map. Although smAP enhances localization performance as shown in \cref{tab:quant_continual_instance}, its lack of a scalable module hinders effective synergy with BPPL.

\begin{table}[h]
    \centering
    
    \vskip -0.08in

    \setlength{\tabcolsep}{2pt}
    \renewcommand{\arraystretch}{1.0}
    
    \resizebox{0.6\columnwidth}{!}{%
        \begin{tabular}{|c|c|c|c|c|}
        \hline
        \textbf{Methods} & RRT-MIL & smAP & \textbf{GDAT} \\
        
        \hline

        \textbf{Params (M)} 
        & 6.00 & 0.60 & \textbf{6.42} \\
        \textbf{VRAM (G)}
        & 10.7 & 3.6 & \textbf{8.8} \\
        \textbf{FLOPs (G)} 
        & 57.4 & 20.6 & \textbf{23.4} \\
        \textbf{Latency (ms)} 
        & 3.36 & 3526 & \textbf{2.58} \\

        \hline
        \end{tabular}
        
    }
    \vskip -0.05in
    \caption{Efficiency analysis of GDAT and OWLoRA.}
    \vskip -0.05in
    \label{tab:scalability_resource_gdat}
\end{table}

\cref{tab:scalability_resource_owlora} compares the memory efficiency of OWLoRA with that of ER and ConSlide. ER and ConSlide are rehearsal-based approaches that require an additional memory buffer, whereas OWLoRA relies on additional learnable parameters. Then, it is evident that OWLoRA is more memory-efficient than the baselines comparing the memory size of their rehearsal buffers to the size of the additional parameters in OWLoRA.

\begin{table}[h]
    \centering
    
    \vskip -0.08in

    \setlength{\tabcolsep}{2pt}
    \renewcommand{\arraystretch}{1.0}
    
    \resizebox{0.75\columnwidth}{!}{%

        \begin{tabular}{|c|c|c|c|}
        \hline
        \textbf{Methods} & ER/30 & ConSlide/30 & \textbf{OWLoRA} \\
        
        \hline  
        \textbf{Params (M)}
        & - & -  & $\textbf{4.6} \times \textbf{10}^{\textbf{-1}}$ \\
        \textbf{Buffer (M)} 
        & $4.5 \times 10^{2}$ & $4.7 \times 10^{2}$ & \textbf{-} \\
        \hline
        \end{tabular}        
    }
    \vskip -0.05in
    \caption{Efficiency analysis of OWLoRA.}
    \vskip -0.05in
    \label{tab:scalability_resource_owlora}
\end{table}

\subsection{Pseudo-label Accuracy of BPPL}

To evaluate the robustness of BPPL in terms of pseudo-label quality, we measured the pseudo-label accuracy for each task during training continual instance classification. From \cref{fig:pl_acc_and_rank_study}(a), BPPL consistently improves the quality of pseudo-labels as training progresses, thereby enhancing localization performance.

\subsection{Low-rank Property of MIL Tasks}

\begin{figure}[t]
\begin{center}

\centerline{\includegraphics[width=1.0\columnwidth]{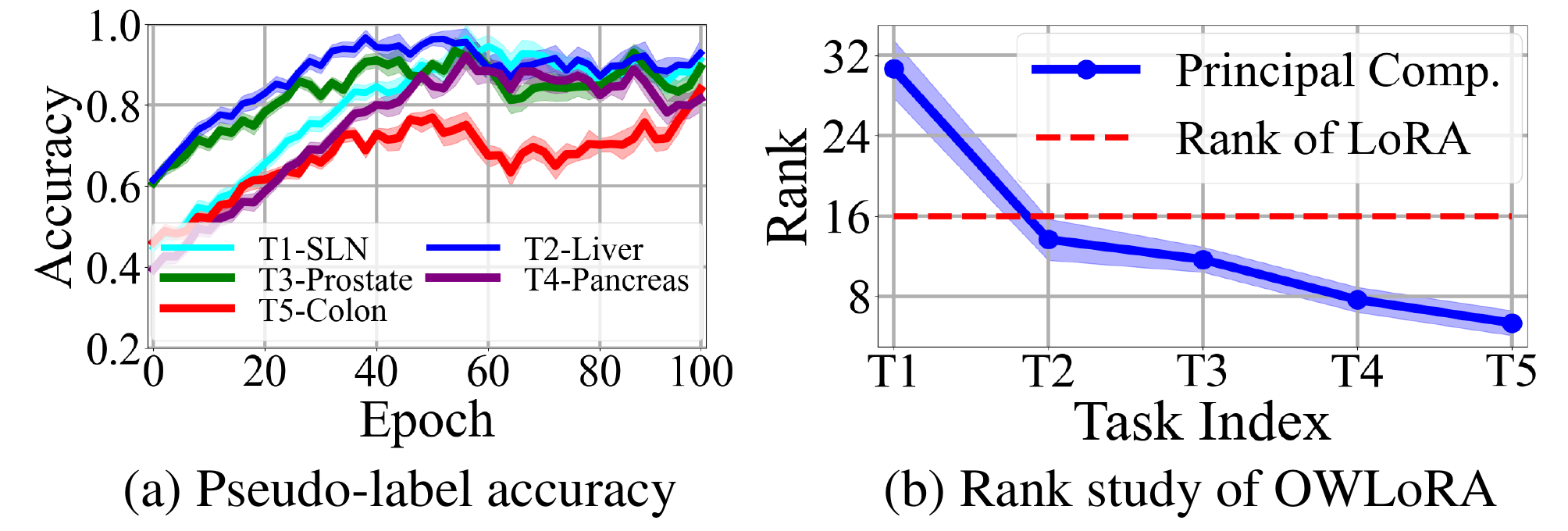}}
\vskip -0.05in
\caption{Pseudo-label accuracy of BPPL and analysis on low-rank property of MIL tasks, supporting the rationale of OWLoRA.}
\label{fig:pl_acc_and_rank_study}
\end{center}
\vskip -0.3in
\end{figure}

Through extensive experiments on continual instance classification, we demonstrated that OWLoRA effectively mitigates forgetting in both bag- and instance-level classification. This is attributed to the low-rank nature of each task and OWLoRA’s orthogonality-based regularization. \cref{fig:pl_acc_and_rank_study}(b) illustrates the number of new singular values required to capture 99\% of the total norm of singular values (across 3 runs), confirming the inherent low-rankedness. Forgetting is further mitigated by enforcing orthogonal subspaces, as supported by \cref{tab:ablation_owlora_eps_d} and \cref{tab:ablation_owlora_supple_1}.

\subsection{Ablation studies on GDAT}

The term $\eta V$ in \cref{eq:cdattn} is introduced to mitigate the reduction in diversity among instance features caused by the grouped attention mechanism. Here, $\eta$ is a hyperparameter.  
\cref{tab:ablation_gdat_eta_supple} shows the performance of continual instance classification on the combined CM-16 and PAIP datasets with different values of $\eta$. We can see that smaller $\eta$ resulted in worse instance-level accuracy  since the reduced diversity among instance features leads to poorer instance discriminability. Meanwhile, when $\eta$ becomes excessively large, the effect of attention-based feature refinement diminishes, ultimately resulting in degraded performance.

\begin{table}[h] 
    \centering

    \setlength{\tabcolsep}{10pt}
    \renewcommand{\arraystretch}{1.05}
    
    \resizebox{\columnwidth}{!}{%
    \begin{tabular}{|c|c|c|c|}
        \hline

        \textbf{$\eta$} &
        \textbf{$\text{ACC}_{\text{inst}}$} ($\uparrow$) & 
        \textbf{$\text{Forget}_{\text{inst}}$} ($\downarrow$) &
        \textbf{$\text{ACC}_{\text{bag}}$} ($\uparrow$) \\
        
        \hline
        
        0 & $68.94 \pm 2.49$  & $14.96 \pm 2.39$  & $62.11 \pm 1.57$  \\
        0.01  & $69.32 \pm 2.74$  & $13.39 \pm 2.52$  & $61.72 \pm 1.92$  \\
        0.05 & $74.13 \pm 2.39$  & $12.94 \pm 2.62$  & $62.84 \pm 1.61$  \\

        \hline
        \textbf{0.1} & $\mathbf{74.15 \pm 1.97}$  & $\mathbf{13.05 \pm 2.31}$  & $\mathbf{62.64 \pm 2.14}$  \\
        \hline
        0.5  & $73.93 \pm 1.72$  & $14.33 \pm 2.10$  & $60.23 \pm 1.75$  \\
        1 & $73.24 \pm 2.19$  & $13.93 \pm 2.38$  & $58.18 \pm 2.30$  \\
        5  & $70.98 \pm 1.89$  & $15.47 \pm 1.66$  & $56.15 \pm 2.08$  \\
        10 & $69.17 \pm 2.09$  & $14.76 \pm 2.39$  & $53.21 \pm 1.82$  \\

        \hline
    \end{tabular}%
    }
    \caption{Ablation studies of the effect of $\eta$ in GDAT (3 runs). The row with \textbf{bold} represents the selected configurations.}

    \label{tab:ablation_gdat_eta_supple}
\end{table}

We investigated the impact of different hyper-parameters on GDAT performance. Relevant hyper-parameters include the grouping window factor $F$ and the number of attention layers in GDAT. When the grouping window factor is set to $F$, the number of instances within a group is $F^2$, and the number of groups is $m=M/F^2$. \cref{tab:ablation_gdat_supple} shows that increasing $F$ deteriorated the performance since unrelated instances could be grouped together and reduce the representation diversity for instances by the attention of GDAT. Meanwhile, increasing the number of attention layers in GDAT resulted in worse performance due to overfitting.

\begin{table}[h] 
    \centering

    \setlength{\tabcolsep}{3pt}
    \renewcommand{\arraystretch}{1.05}
    
    \resizebox{\columnwidth}{!}{%
    \begin{tabular}{|c|c|c|c|c|}
        \hline

        \textbf{F} &
        \textbf{\# of layers} & 
        \textbf{$\text{ACC}_{\text{inst}}$} ($\uparrow$) & 
        \textbf{$\text{Forget}_{\text{inst}}$} ($\downarrow$) &
        \textbf{$\text{ACC}_{\text{bag}}$} ($\uparrow$) \\
        
        \hline
        
        8  & 2  & $74.20 \pm 2.24$  & $14.23 \pm 2.04$  & $62.92 \pm 2.52$  \\
        32 & 2  & $72.25 \pm 2.56$  & $14.57 \pm 2.66$  & $61.87 \pm 2.40$  \\
        64 & 2  & $68.23 \pm 3.23$  & $16.23 \pm 2.43$  & $56.24 \pm 3.24$  \\

        \hline
        \textbf{16} & \textbf{2} & $\mathbf{74.15 \pm 1.97}$  & $\mathbf{13.05 \pm 2.31}$  & $\mathbf{62.64 \pm 2.14}$  \\
        \hline
        16 & 1  & $72.43 \pm 2.39$  & $14.54 \pm 2.46$  & $60.83 \pm 2.32$  \\
        16 & 4  & $74.45 \pm 2.08$  & $13.33 \pm 1.94$  & $62.82 \pm 1.72$  \\
        16 & 8  & $73.20 \pm 2.55$  & $13.92 \pm 2.72$  & $61.33 \pm 2.40$  \\
        16 & 12 & $72.01 \pm 2.04$  & $14.90 \pm 2.44$  & $60.89 \pm 2.06$  \\

        \hline
    \end{tabular}%
    }
    \caption{Ablation studies of the effect of grouping window factor $F$ and the number of attention layers in GDAT (3 runs). The row with \textbf{bold} represents the selected configurations.}

    \label{tab:ablation_gdat_supple}
\end{table}

\subsection{Ablation studies on BPPL}
We investigated the impact of different hyper-parameters on BPPL performance. Relevant hyper-parameters include temperature $T$ and coefficient $\tau$. \cref{tab:ablation_bppl_supple_1} shows that increasing $T$ too much deteriorates the performance due to small probability values, while decreasing $T$ too much deteriorates the performance due to large probability values. Additionally, increasing $\tau$ too much deteriorated the performance due to strict pseudo-labeling, while decreasing $\tau$ deteriorated the performance due to inaccurate pseudo-labels.

\begin{table}[h] 
    \centering

    \setlength{\tabcolsep}{5pt}
    \renewcommand{\arraystretch}{1.05}
    
    \resizebox{\columnwidth}{!}{%
    \begin{tabular}{|c|c|c|c|c|}
        \hline

        T & 
        $\mathbf{\tau}$ & 
        \textbf{$\text{ACC}_{\text{inst}}$} ($\uparrow$) & 
        \textbf{$\text{Forget}_{\text{inst}}$} ($\downarrow$) &
        \textbf{$\text{ACC}_{\text{bag}}$} ($\uparrow$) \\
        
        \hline
        
        0.01 & 0.35 & $60.32 \pm 3.52$ & $13.02 \pm 3.09$ & $60.92 \pm 2.10$ \\
        0.05 & 0.35 & $68.23 \pm 2.69$ & $12.43 \pm 2.72$ & $61.52 \pm 1.93$ \\
        0.1  & 0.35 & $73.23 \pm 2.46$ & $13.49 \pm 2.42$ & $62.04 \pm 2.01$ \\
        1    & 0.35 & $73.12 \pm 2.98$ & $13.92 \pm 2.19$ & $62.40 \pm 1.92$ \\
        5    & 0.35 & $62.23 \pm 2.82$ & $14.92 \pm 2.39$ & $61.24 \pm 1.82$ \\
        \hline
        \textbf{0.5} & \textbf{0.35} & $\mathbf{74.15 \pm 1.97}$ & $\mathbf{13.05 \pm 2.31}$ & $\mathbf{62.64 \pm 2.14}$ \\
        \hline
        0.5  & 0.05 & $61.23 \pm 3.22$ & $15.12 \pm 2.72$ & $61.23 \pm 2.92$ \\
        0.5  & 0.1  & $64.23 \pm 2.32$ & $17.23 \pm 1.98$ & $61.63 \pm 1.62$ \\
        0.5  & 0.25 & $70.37 \pm 2.23$ & $14.62 \pm 2.93$ & $61.73 \pm 1.40$ \\
        0.5  & 0.45 & $71.24 \pm 2.42$ & $14.34 \pm 3.09$ & $63.20 \pm 2.20$ \\

        \hline
    \end{tabular}%
    }
    \caption{Ablation studies of the effect of $T$ and $\tau$ in BPPL (3 runs). The row with \textbf{bold} represents the selected configurations.}

    \label{tab:ablation_bppl_supple_1}
\end{table}

We also investigated the impact of additional hyper-parameters, including the regularization coefficients $\lambda_1$ and $\lambda_2$. \cref{tab:ablation_bppl_supple_2} shows that increasing $\lambda_1$ too much deteriorated the performance due to placing too much weight on instance localization, while decreasing $\lambda_1$ deteriorated the performance on instance classification due to less weight on training for localization. Furthermore, increasing $\lambda_2$ too much deteriorated the performance due to excessive bag separation, while decreasing $\lambda_2$ deteriorated the performance due to insufficient bag separation.

\begin{table}[h] 
    \centering

    \setlength{\tabcolsep}{8pt}
    \renewcommand{\arraystretch}{1.05}
    
    \resizebox{\columnwidth}{!}{%
    \begin{tabular}{|c|c|c|c|c|}
        \hline

        $\mathbf{\lambda_1}$ & 
        $\mathbf{\lambda_2}$ & 
        \textbf{$\text{ACC}_{\text{inst}}$} ($\uparrow$) & 
        \textbf{$\text{Forget}_{\text{inst}}$} ($\downarrow$) &
        \textbf{$\text{ACC}_{\text{bag}}$} ($\uparrow$) \\
        
        \hline
        
        0.1 & 1  & $72.32 \pm 2.33$ & $13.89 \pm 2.87$ & $62.47 \pm 1.98$ \\
        1   & 1    & $73.84 \pm 2.93$ & $13.24 \pm 2.52$ & $62.92 \pm 2.52$ \\
        5   & 1    & $67.23 \pm 1.92$ & $11.23 \pm 1.52$ & $57.24 \pm 1.62$ \\
        10  & 1    & $61.23 \pm 2.32$ & $9.23 \pm 2.24$  & $54.23 \pm 2.20$ \\
        \hline
        \textbf{0.5}  & \textbf{1}  & $\mathbf{74.15 \pm 1.97}$ & $\mathbf{13.05 \pm 2.31}$ & $\mathbf{62.64 \pm 2.14}$ \\
        \hline
        0.5 & 0.1  & $70.92 \pm 2.60$ & $17.34 \pm 2.72$ & $62.82 \pm 2.12$ \\
        0.5 & 0.3  & $71.72 \pm 2.72$ & $15.94 \pm 2.11$ & $62.52 \pm 1.82$ \\
        0.5 & 3    & $70.24 \pm 1.82$ & $16.68 \pm 1.43$ & $61.24 \pm 2.67$ \\
        0.5 & 10   & $62.23 \pm 2.53$ & $12.35 \pm 3.19$ & $55.23 \pm 3.06$ \\

        \hline
    \end{tabular}%
    }
    \caption{Ablation studies of the effect of $\lambda_1$ and $\lambda_2$ in BPPL (3 runs). The row with \textbf{bold} represents the selected configurations.}

    \label{tab:ablation_bppl_supple_2}
\end{table}

\subsection{Ablation studies on OWLoRA}

We  investigated the impact of $\epsilon$ for first task and the rank $d$ for subsequent tasks in OWLoRA. \cref{tab:ablation_owlora_eps_d} shows decreasing $\epsilon$ deteriorated the performance on instance classification. It is attributed to the forgetting of the first task. Meanwhile, decreasing the rank $d$ resulted in worse instance classification by the reduced adaptability to new tasks. 


\begin{table}[h]
    \centering

    \setlength{\tabcolsep}{9pt}
    \renewcommand{\arraystretch}{1.05}
    
    \resizebox{\columnwidth}{!}{%
    \begin{tabular}{|c|c|c|c|c|}
        \hline

        \textbf{$\epsilon$} & \textbf{$d$} & 
        \textbf{$\text{ACC}_{\text{inst}}$} ($\uparrow$) & 
        \textbf{$\text{Forget}_{\text{inst}}$} ($\downarrow$) &
        \textbf{$\text{ACC}_{\text{bag}}$} ($\uparrow$) \\
        
        \hline
        0.5 & 16
        & $68.23 \pm 2.30$ & $19.47 \pm 2.73$ & $56.10 \pm 2.49$ \\
        
        0.7 & 16
        & $70.42 \pm 2.29$ & $16.78 \pm 2.19$ & $59.35 \pm 2.44$ \\
        
        0.9 & 16
        & $71.23 \pm 2.49$ & $15.04 \pm 2.59$ & $61.23 \pm 2.87$ \\

        0.999 & 16
        & $74.22 \pm 2.05$ & $12.99 \pm 2.10$ & $62.24 \pm 2.19$ \\

        \hline 
        \textbf{0.99} & \textbf{16}
        & \textbf{74.15} $\pm$ \textbf{1.97} & \textbf{13.05} $\pm$ \textbf{2.31} & \textbf{62.64} $\pm$ \textbf{2.14} \\
        
        \hline
        
        0.99 & 2
        & $67.23 \pm 1.88$ & $20.44 \pm 1.84$ & $58.38 \pm 1.61$ \\
        
        0.99 & 4
        & $70.04 \pm 1.71$ & $17.05 \pm 1.79$ & $59.92 \pm 2.04$ \\

        0.99 & 8
        & $72.86 \pm 2.65$ & $14.98 \pm 1.94$ & $61.04 \pm 1.98$ \\

        0.99 & 32
        & $74.19 \pm 2.04$ & $13.37 \pm 2.46$ & $63.33 \pm 2.46$ \\
        
        \hline
    \end{tabular}%
    }
    \caption{Ablation studies of the impact of $\epsilon$ for first task and $d$ for subsequent tasks in OWLoRA for continual instance classification (3 runs). The row with \textbf{bold} represents the selected configurations.}

    \label{tab:ablation_owlora_eps_d}

    \vskip -0.05in
    
\end{table}

We further investigated the impact of the hyper-parameter $\lambda_3$ on the performance of OWLoRA. \cref{tab:ablation_owlora_supple_1} shows that increasing $\lambda_3$ deteriorated the performance because it excessively enforces the orthogonality on the basis for mitigating forgetting, which hinders to learn new tasks. Meanwhile, decreasing $\lambda_3$ too much also deteriorated the performance because of the inability to mitigate forgetting.

\begin{table}[h] 
    \centering

    \setlength{\tabcolsep}{8pt}
    \renewcommand{\arraystretch}{1.05}
    
    \resizebox{\columnwidth}{!}{%
    \begin{tabular}{|c|c|c|c|}
        \hline

        $\mathbf{\lambda_3}$ & 
        \textbf{$\text{ACC}_{\text{inst}}$} ($\uparrow$) & 
        \textbf{$\text{Forget}_{\text{inst}}$} ($\downarrow$) &
        \textbf{$\text{ACC}_{\text{bag}}$} ($\uparrow$) \\
        
        \hline
        
        0.1  & $70.32 \pm 2.91$  & $17.50 \pm 2.72$  & $55.33 \pm 3.24$  \\
        0.5  & $72.24 \pm 2.33$  & $14.88 \pm 2.29$  & $60.23 \pm 2.57$  \\
        \hline
        \textbf{1}  & $\mathbf{74.15 \pm 1.97}$  & $\mathbf{13.05 \pm 2.31}$  & $\mathbf{62.64 \pm 2.14}$  \\
        \hline
        5    & $74.21 \pm 2.04$  & $13.08 \pm 2.46$  & $61.94 \pm 2.22$  \\
        10   & $71.39 \pm 1.64$  & $16.42 \pm 1.82$  & $60.92 \pm 1.50$  \\

        \hline
    \end{tabular}%
    }
    \caption{Ablation studies of the effect of $\lambda_3$ in OWLoRA (3 runs). The row with \textbf{bold} represents the selected configurations.}

    \label{tab:ablation_owlora_supple_1}
\end{table}



\subsection{Additional Structural Ablation Studies}

We further performed additional structural ablation studies to investigate the design choices of GDAT and BPPL. From \cref{tab:structure_ablation_gdat}, GDAT (double attention) achieves competitive performance to single attention with better scalability. From \cref{tab:structure_ablation_bppl}, BPPL fails to learn discriminative prototypes without $\mathcal{L}_{\text{sep}}$, reducing performance. Since three filtering factors in BPPL are crucial for high-quality pseudo-labels, removing any of them leads to degradation in localization.

\begin{table}[h]
    \centering
    
    \setlength{\tabcolsep}{2pt}
    \renewcommand{\arraystretch}{1.0}

    \resizebox{\columnwidth}{!}{%
    \begin{tabular}{|c|c|c|c|c|}
        \hline
        
        \textbf{Model} 
        & \textbf{$\text{ACC}_{\text{bag}}$ ($\uparrow$)} 
        & \textbf{$\text{ACC}_{\text{inst}}$ ($\uparrow$)} 
        & \textbf{IoU ($\uparrow$)} 
        & \textbf{Dice ($\uparrow$)}
        \\
        
        \hline
        
        Single Attn
        & $73.53 \pm 1.64$ 
        & $80.87 \pm 1.82$ 
        & $50.69 \pm 2.74$ 
        & $62.32 \pm 2.35$
        \\

        \textbf{GDAT}
        & $\mathbf{72.94 \pm 1.28}$ 
        & $\mathbf{80.55 \pm 2.34}$ 
        & $\mathbf{50.35 \pm 3.43}$ 
        & $\mathbf{61.70 \pm 2.87}$
        \\

        \hline
    \end{tabular}

    }
    \vskip -0.1in
    \caption{Comparison of single attention and GDAT (double attention) when \textbf{combined with BPPL} on merged dataset (3 runs).}
    \vskip -0.05in
    \label{tab:structure_ablation_gdat}
\end{table}

\begin{table}[h]
    \centering

    \setlength{\tabcolsep}{3pt}
    \renewcommand{\arraystretch}{1.0}
    
    \resizebox{\columnwidth}{!}{%
    \begin{tabular}{|c|c|c|c|c|}
        \hline
        
        \textbf{Ablating Comp.} 
        & \textbf{$\text{ACC}_{\text{inst}}$ ($\uparrow$)} & \textbf{$\text{Forget}_{\text{inst}}$ ($\downarrow$)} & \textbf{Dice} ($\uparrow$) & \textbf{ $\text{ACC}_{\text{bag}}$ ($\uparrow$)} \\

        \hline

        \textbf{CoMEL} 
        & $\textbf{74.15} \pm \textbf{1.97}$ 
        & $\textbf{13.05} \pm \textbf{2.31}$
        & $\textbf{52.27} \pm \textbf{2.42}$ 
        & $\textbf{62.64} \pm \textbf{2.14}$ \\

        w/o $\mathcal{L}_{\text{sep}}$
        & $69.82 \pm 2.29$ & $17.86 \pm 2.43$
        & $45.60 \pm 2.56$  & $62.84 \pm 2.18$ \\

        w/o $\mathbbm{1} (\hat{Y} = Y)$
        & $66.2 \pm 2.1$ & $15.0 \pm 1.9$
        & $31.6 \pm 2.4$ & $60.4 \pm 1.6$ \\

        w/o $\mathbbm{1} (\hat{p}_m > \tau_1)$
        & $69.28 \pm 1.76$ & $15.84 \pm 1.96$
        & $37.51 \pm 2.65$ & $61.40 \pm 1.66$ \\

        w/o $\mathbbm{1}(Y_{\text{bag}} \in \mathcal{Y}_{\text{pos}})$
        & $66.54 \pm 1.72$ & $15.81 \pm 1.73$
        & $34.77 \pm 2.23$ & $60.89 \pm 2.32$ \\
        
        \hline
    \end{tabular}%
    }
    \vskip -0.1in

    \caption{Ablation studies of loss components and filtering factors in BPPL on continual instance classification (3 runs).}
    \vskip -0.1in
    \label{tab:structure_ablation_bppl}
\end{table}


\begin{table*}[t]
    \centering
    
    \setlength{\tabcolsep}{8pt}
    \renewcommand{\arraystretch}{1.0}
    
    \resizebox{\textwidth}{!}{%
    \begin{tabular}{|c|c|c|c|c|c|c|}
        \hline
        \textbf{IL Type} 
        & \textbf{Method} 
        & \textbf{$\text{ACC}_{\text{inst}}$ ($\uparrow$)} 
        & \textbf{$\text{Forget}_{\text{inst}}$ ($\downarrow$)} 
        & \textbf{IoU ($\uparrow$)} 
        & \textbf{Dice ($\uparrow$)} 
        & \textbf{$\text{ACC}_{\text{bag}}$ ($\uparrow$)} \\
        \hline
    
        \multirow{2}{*}{Upper Bound} 
        & Joint (Full label) 
        & $90.32 \pm 3.34$ 
        & - 
        & $67.72 \pm 2.04$ 
        & $77.23 \pm 2.39$ 
        & $75.97 \pm 3.71$ \\
        
        & Joint (Weak label) 
        & $79.50 \pm 2.72$ 
        & - 
        & $51.67 \pm 2.19$ 
        & $61.09 \pm 2.64$ 
        & $72.28 \pm 3.49$ \\
        
        \hline
        Lower Bound 
        & Finetune 
        & $56.00 \pm 2.17$ 
        & $29.12 \pm 3.37$ 
        & $10.82 \pm 2.30$ 
        & $18.71 \pm 2.39$ 
        & $9.42 \pm 3.58$ \\
    
        \hline
        \multirow{2}{*}{Regularization-based} 
        & EWC 
        & $60.65 \pm 2.88$ 
        & $23.79 \pm 2.37$ 
        & $13.81 \pm 3.00$ 
        & $20.75 \pm 2.52$ 
        & $15.37 \pm 2.79$ \\
        
        & LwF 
        & $61.83 \pm 2.87$ 
        & $24.30 \pm 2.26$ 
        & $16.29 \pm 2.92$ 
        & $23.96 \pm 3.19$ 
        & $18.82 \pm 3.23$ \\
    
        \hline
        \multirow{6}{*}{Rehearsal-based} 
        & A-GEM/30 
        & $63.19 \pm 3.33$ 
        & $22.78 \pm 3.39$ 
        & $18.48 \pm 2.61$ 
        & $25.23 \pm 3.64$ 
        & $38.65 \pm 2.57$ \\
        
        & ER/30 
        & $63.27 \pm 2.83$ 
        & $23.01 \pm 3.72$ 
        & $17.16 \pm 3.11$ 
        & $24.80 \pm 3.08$ 
        & $43.70 \pm 2.31$ \\
        
        & ER/100 
        & $67.09 \pm 3.28$ 
        & \underline{$19.49 \pm 2.84$} 
        & $21.29 \pm 3.02$ 
        & $29.12 \pm 3.30$ 
        & $44.82 \pm 3.43$ \\
        
        & DER++/30 
        & $65.83 \pm 2.74$ 
        & $21.76 \pm 3.00$ 
        & $21.03 \pm 3.11$ 
        & $28.06 \pm 2.52$ 
        & $45.79 \pm 3.02$ \\
        
        & ER-ACE/30 
        & $66.64 \pm 3.58$ 
        & $20.84 \pm 3.10$ 
        & $17.64 \pm 3.40$ 
        & $24.83 \pm 3.19$ 
        & $48.02 \pm 3.63$ \\
        
        & ConSlide/30 
        & $65.52 \pm 3.12$ 
        & $21.27 \pm 3.13$ 
        & $19.97 \pm 2.56$ 
        & $25.67 \pm 3.60$ 
        & $52.92 \pm 3.29$ \\
    
        \hline

        Prompt-tuning-based
        & QPMIL-VL \cite{gou2024queryable} 
        & $66.37 \pm 2.21$ 
        & $20.51 \pm 2.17$
        & $25.15 \pm 2.35$ 
        & $34.28 \pm 2.19$ 
        & \underline{$54.83 \pm 2.39$} \\
        
        \hline

        \multirow{3}{*}{LoRA-based} 
        & LoRA finetune 
        & $61.27 \pm 3.03$ 
        & $22.95 \pm 2.45$ 
        & $21.44 \pm 3.02$ 
        & $28.65 \pm 3.72$ 
        & $28.01 \pm 3.45$ \\
        
        & InfLoRA 
        & \underline{$67.60 \pm 2.84$} 
        & $20.37 \pm 3.65$ 
        & \underline{$30.33 \pm 3.42$} 
        & \underline{$39.54 \pm 3.14$} 
        & $51.53 \pm 2.61$ \\
        
        & \textbf{CoMEL (Ours)} 
        & $\mathbf{70.65} \pm \mathbf{2.84}$ 
        & $\mathbf{16.93} \pm \mathbf{2.79}$ 
        & $\mathbf{38.81} \pm \mathbf{3.28}$ 
        & $\mathbf{48.22} \pm \mathbf{3.69}$ 
        & $\mathbf{57.72} \pm \mathbf{3.26}$ \\
        \hline
    \end{tabular}
    }
    
    \vskip -0.1in
    
    \caption{Additional quantitative results of CL methods on instance classification in the reversed continual MIL setup. The best and second best results are marked as \textbf{bold} and \underline{underline}. Each experiment consisted of 10 runs. The experiments were conducted on five sequential organ datasets from combined CM-16 and PAIP. For baselines, we applied the CL approaches upon our GDAT+BPPL, except for ConSlide. All metrics are reported in percentages. CoMEL achieved the highest performance across all metrics while minimizing the forgetting.}
    
    \label{tab:additional_quantitative_supple_1}

    \vskip -0.02in
    
\end{table*}

\begin{table*}[t]
    \centering

    \setlength{\tabcolsep}{11pt}
    \renewcommand{\arraystretch}{1.0}
    
    \resizebox{\textwidth}{!}{%
    \begin{tabular}{|c|c|c|c|c|c|c|}
        \hline
        
        \textbf{Model} 
        & \textbf{$\text{ACC}_{\text{bag}}$ ($\uparrow$)} 
        & \textbf{$\text{AUC}_{\text{bag}}$ ($\uparrow$)} 
        & \textbf{$\text{F1}_{\text{bag}}$ ($\uparrow$)}
        & \textbf{$\text{ACC}_{\text{inst}}$ ($\uparrow$)}
        & \textbf{IoU ($\uparrow$)} 
        & \textbf{Dice ($\uparrow$)} \\
        
        \hline
        
        ABMIL \cite{lu2021data} 
        & $91.13 \pm 1.34$ & $94.21 \pm 1.07$ & $86.42 \pm 1.39$
        & $77.53 \pm 2.28$ 
        & $38.22 \pm 3.01$ & $49.33 \pm 2.71$ \\
        
        DS-MIL \cite{li2021dual}
        & $90.91 \pm 1.59$ & $93.83 \pm 1.24$ & $86.11 \pm 1.48$
        & $72.18 \pm 2.17$ 
        & $28.59 \pm 2.64$ & $37.81 \pm 2.42$ \\
        
        TransMIL \cite{shao2021transmil}
        & $91.51 \pm 1.33$ & $94.02 \pm 1.19$ & $87.25 \pm 1.36$
        & $76.03 \pm 2.51$
        & $35.11 \pm 2.41$ & $45.08 \pm 2.29$ \\

        RRT-MIL \cite{tang2024feature} 
        & $\underline{93.23 \pm 1.08}$ & $\underline{96.14 \pm 1.13}$ & $89.53 \pm 0.94$
        & $73.44 \pm 1.98$
        & $32.14 \pm 1.97$ & $41.01 \pm 2.14$ \\

        smAP \cite{castrosm} 
        & $91.42 \pm 1.21$ & $94.06 \pm 1.38$ & $87.63 \pm 1.31$
        & $\underline{86.01 \pm 1.95}$
        & $\underline{47.12 \pm 2.51}$ & $\underline{59.11 \pm 2.67}$ \\

        \hline

        \textbf{GDAT (Ours)} 
        & $\mathbf{93.21 \pm 1.31}$ & $\mathbf{96.56 \pm 1.14}$ & $\mathbf{90.02 \pm 0.97}$
        & $78.98 \pm 2.36$ 
        & $39.91 \pm 2.74$ & $51.61 \pm 2.53$ \\

        \textbf{GDAT+BPPL (Ours)} 
        & $93.04 \pm 1.42$ & $95.74 \pm 1.25$ & $\underline{89.62 \pm 1.27}$
        & $\mathbf{87.73 \pm 2.61}$ 
        & $\mathbf{53.21 \pm 3.04}$ & $\mathbf{64.17 \pm 3.23}$ \\
        
        \hline
    \end{tabular}
    }
    \vskip -0.1in

    \caption{Comparison of different MIL models on a single dataset CM-16. We evaluated bag classification using ACC$_{\text{bag}}$, AUC$_{\text{bag}}$, and F1$_{\text{bag}}$.
    For instance classification, We evaluated using ACC$_{\text{inst}}$, IoU, and Dice score. The best and second-best results are marked as \textbf{bold} and \underline{underline}, respectively. Each experiment consisted of 10 runs. Our proposed method achieved the best performance across all metrics, demonstrating its superiority in instance-level classification.}
    \label{tab:further_results_single_1}

    \vskip -0.1in
    
\end{table*}

\begin{table}[h]
    \centering

    \setlength{\tabcolsep}{3pt}
    \renewcommand{\arraystretch}{1.1}
    
    \resizebox{\columnwidth}{!}{%
    \begin{tabular}{|c|c|c|c|}
        \hline
        
        \textbf{Method} & \textbf{$\text{ACC}_{\text{bag}}$ ($\uparrow$)} & \textbf{$\text{Forget}_{\text{bag}}$ ($\downarrow$)} &  \textbf{M.$\text{ACC}_{\text{bag}}$ ($\uparrow$)} \\
        \hline

        Joint
        & $91.18 \pm 2.31$ & -  & $93.42 \pm 2.53$ \\

        \hline
        
        Finetune
        & $24.41 \pm 3.74$ & $66.13 \pm 3.87$ & $79.36 \pm 3.43$ \\

        \hline
        
        EWC
        & $25.02 \pm 4.11$ & $64.58 \pm 3.98$  & $84.02 \pm 3.66$ \\

        LwF
        & $26.61 \pm 3.73$ & $62.35 \pm 3.40$ & $87.91 \pm 3.94$ \\
        
        \hline
        
        A-GEM/30
        & $45.62 \pm 4.03$ & $48.42 \pm 4.14$ & $87.44 \pm 4.38$ \\
        
        ER/30
        & $67.79 \pm 2.94$ & $25.48 \pm 3.07$ & $89.33 \pm 5.28$ \\

        ER/100
        & $70.66 \pm 2.83$ & $23.61 \pm 2.98$ & $90.21 \pm 3.38$ \\
        
        DER++/30
        & $68.94 \pm 4.02$ & $24.11 \pm 4.23$ & $89.76 \pm 3.97$ \\
        
        ConSlide/30
        & $76.21 \pm 2.93$ & $17.66 \pm 3.13$ & $90.01 \pm 3.24$ \\

        \hline

        QPMIL-VL 
        & $79.92 \pm 3.54$ & $\underline{13.24 \pm 3.28}$ & $\underline{90.80 \pm 2.95}$ \\

        \hline

        LoRA finetune
        & $45.89 \pm 4.22$ & $38.52 \pm 3.80$ & $87.07 \pm 3.21$ \\
        
        InfLoRA
        & $\underline{80.31 \pm 4.14}$ & $13.59 \pm 3.64$ & $88.46 \pm 3.72$ \\

        \textbf{CoMEL (Ours)} 
        & $\mathbf{83.27 \pm 3.63}$ & $\mathbf{11.51 \pm 3.78}$ & $\mathbf{91.42 \pm 3.17}$ \\
        
        \hline
    \end{tabular}
    }
    \vskip -0.1in
    \caption{Additional comparison of methods for slide-level classification on the TCGA dataset with reversed task order. The best and second-best results are marked as \textbf{bold} and \underline{underline}. Each experiment consisted of 10 runs.}
    \label{tab:additional_quantitative_supple_2}
    \vskip -0.05in
\end{table}

\section{Additional Quantitative Results}

\subsection{\textbf{Results on Continual Instance Classification}}  \label{sec:inst_reversed}

\cref{tab:additional_quantitative_supple_1} presents additional quantitative results for tumor detection across the sequential datasets of combined CM16 and PAIP, but the sequence is reversed from the experiment in the main text. That is, the sequence is: colon, pancreas, prostate, liver, and sentinel lymph nodes.
From the results for ACC$_{\text{bag}}$ and ACC$_{\text{inst}}$, we observed that while rehearsal-based approaches mitigated forgetting in bag classification with better performance than regularization-based approaches, they still suffered from substantial forgetting in instance classification similar to our experiment in the main text. On the other hand, InfLoRA outperformed them in both ACC$_{\text{inst}}$ and Forget$_{\text{inst}}$ with competitive performance in terms of ACC$_{\text{bag}}$ also similar to our initial experiment. Upon the LoRA-based CL approach, our CoMEL achieved the best performance across all metrics. In particular, CoMEL outperformed in terms of IoU and Dice by a large margin, demonstrating its effectiveness in preserving localization in the continual MIL even when the sequence is reversed.

\subsection{Results on Continual Bag Classification} \label{sec:bag_reversed}
To further evaluate the performance on continual bag classification, we compared CoMEL against the same baselines in the continual instance classification, but with the sequence reversed. That is, the sequence of organ for continual tasks is: ESCA, RCC, BRCA, and NSCLC.  \cref{tab:additional_quantitative_supple_2} illustrates the performance of CL baselines and CoMEL for continual bag classification on the TCGA datasets with reverse sequence. Just like in the initial experiment, CoMEL achieved the highest performance in terms of ACC\textsubscript{bag} and Forget\textsubscript{bag}, demonstrating its strong performance while effectively preserving previously learned knowledge. Consistently, CoMEL also achieved the highest masked bag-level accuracy (M.ACC\textsubscript{bag}) compared to the baselines. Rehearsal-based approaches such as ER and ConSlide demonstrated their effectiveness in mitigating catastrophic forgetting for continual bag classification, while regularization-based methods suffered from severe forgetting.

\subsection{\textbf{Results on Single Dataset}} \label{sec:single}
We further evaluated our method's bag and instance classification ability on the single dataset CM-16. From \cref{tab:further_results_single_1}, we can see that GDAT achieved the best performance on bag classification compared to the baselines, demonstrating its effectiveness.
Furthermore, the removal of BPPL from GDAT results in a notable performance drop in instance classification for localization. This indicates that BPPL is an important component for effective instance localization.

\subsection{\textbf{Additional Metrics with Various Backbones}} \label{sec:supp_var_backbones}

\cref{tab:missing_metrics} presents the results of additional metrics that were omitted from \cref{tab:quant_extractors}. It demonstrates that CoMEL performs well across various backbones for the missing metrics.

\begin{table}[h]
    \centering

    \setlength{\tabcolsep}{3pt}
    \renewcommand{\arraystretch}{1.0}
    
    \resizebox{\columnwidth}{!}{
    \begin{tabular}{|c|c|c|c|c|}
        \hline
        
        \textbf{Model} & ResNet50 & PLIP & CONCH & UNI \\
        
        \hline

        \textbf{IoU ($\uparrow$)} 
        & $36.19 \pm 2.24$ 
        & $39.34 \pm 1.87$ 
        & $37.16 \pm 2.31$ 
        & $42.78 \pm 1.93$ \\
        
        \textbf{$\text{ACC}_{\text{bag}}$ ($\uparrow$)}
        & $55.29 \pm 1.47$ 
        & $59.22 \pm 2.04$ 
        & $54.20 \pm 1.49$ 
        & $61.59 \pm 2.11$ \\
        
        \textbf{$\text{AUC}_{\text{bag}}$ ($\uparrow$)} 
        & $88.86 \pm 1.21$ 
        & $89.38 \pm 1.03$ 
        & $88.27 \pm 1.86$ 
        & $90.22 \pm 1.74$\\
        
        \hline
    \end{tabular}
    }

    \vskip -0.1in

    \caption{Additional metrics with different feature extractors on continual instance classification.}
    \vskip -0.05in
    \label{tab:missing_metrics}
    
\end{table}

\section{Additional Qualitative Results} \label{sec:quali_additional}
\begin{figure*}[t]
\begin{center}
\centerline{\includegraphics[width=1.0\textwidth]{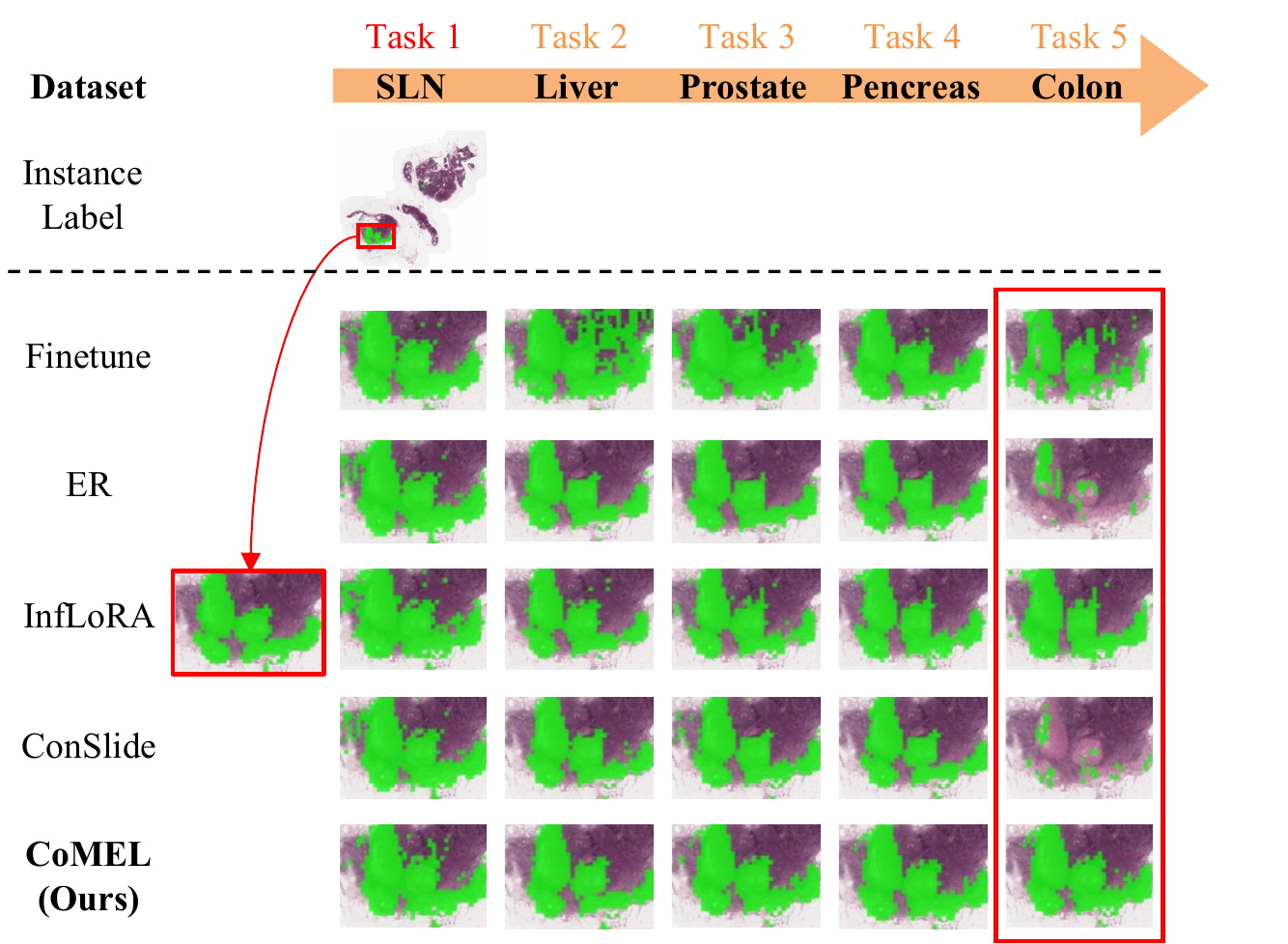}}
 \vskip -0.05in
\caption{Additional qualitative results of localization across sequential organ datasets under the continual MIL setup. Each column is the localization performance on Task 1 as the learned organ changes over sequential tasks. Each row corresponds to CL methods including CoMEL. CoMEL successfully preserved the localization quality across all tasks, while baselines increase false positives or false negatives.}
\label{fig:add_quali_instance_1}
\vskip -0.4in
\end{center}
\end{figure*}

\begin{figure*}[t]
\begin{center}
\centerline{\includegraphics[width=1.0\textwidth]{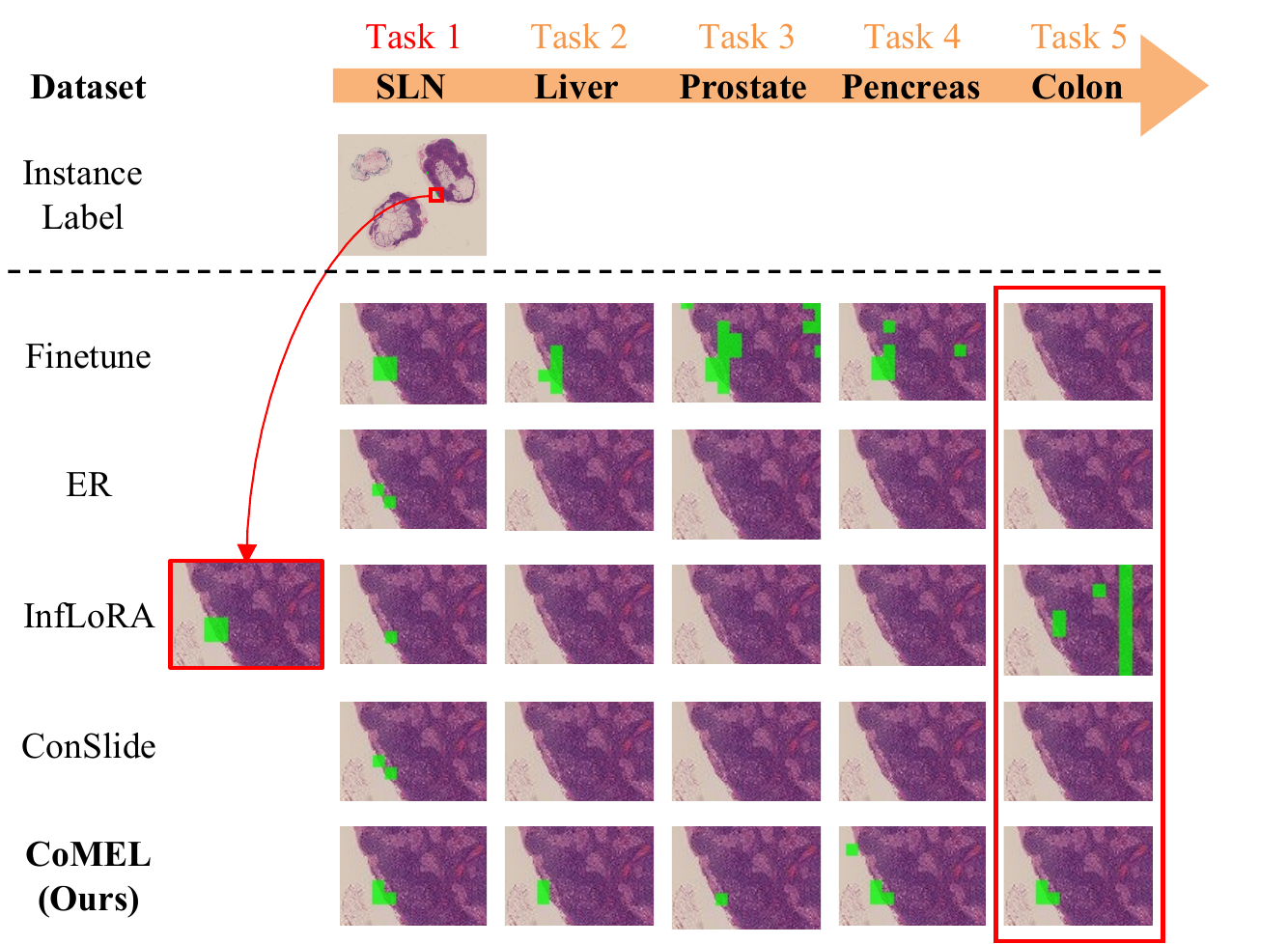}}
 \vskip -0.05in
\caption{Additional qualitative results of localization across sequential organ datasets under the continual MIL setup. Each column is the localization performance on Task 1 as the learned organ changes over sequential tasks. Each row corresponds to CL methods including CoMEL. CoMEL successfully preserved the localization quality across all tasks, while baselines increase false positives or false negatives.}
\label{fig:add_quali_instance_2}
\vskip -0.4in
\end{center}
\end{figure*}

\begin{figure*}[t]
\begin{center}
\centerline{\includegraphics[width=0.9\textwidth]{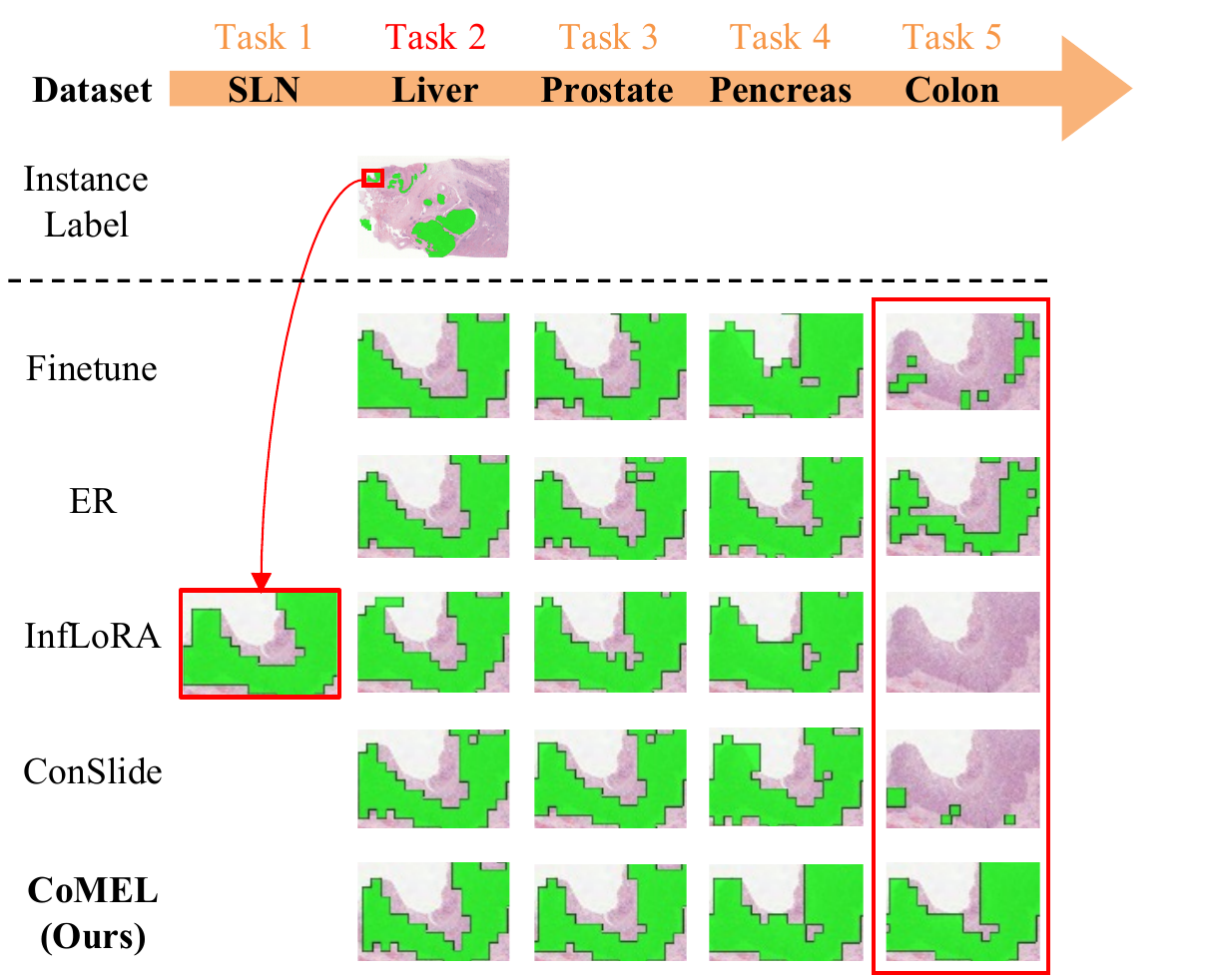}}
 \vskip -0.05in
\caption{Additional qualitative results of localization across sequential organ datasets under the continual MIL setup. Each column is the localization performance on Task 2 as the learned organ changes over sequential tasks. Each row corresponds to CL methods including CoMEL. CoMEL successfully preserved the localization quality across all tasks, while baselines increase false positives or false negatives.}
\label{fig:add_quali_instance_3}
\vskip -0.4in
\end{center}
\end{figure*}

\begin{figure*}[t]
\begin{center}
\centerline{\includegraphics[width=0.9\textwidth]{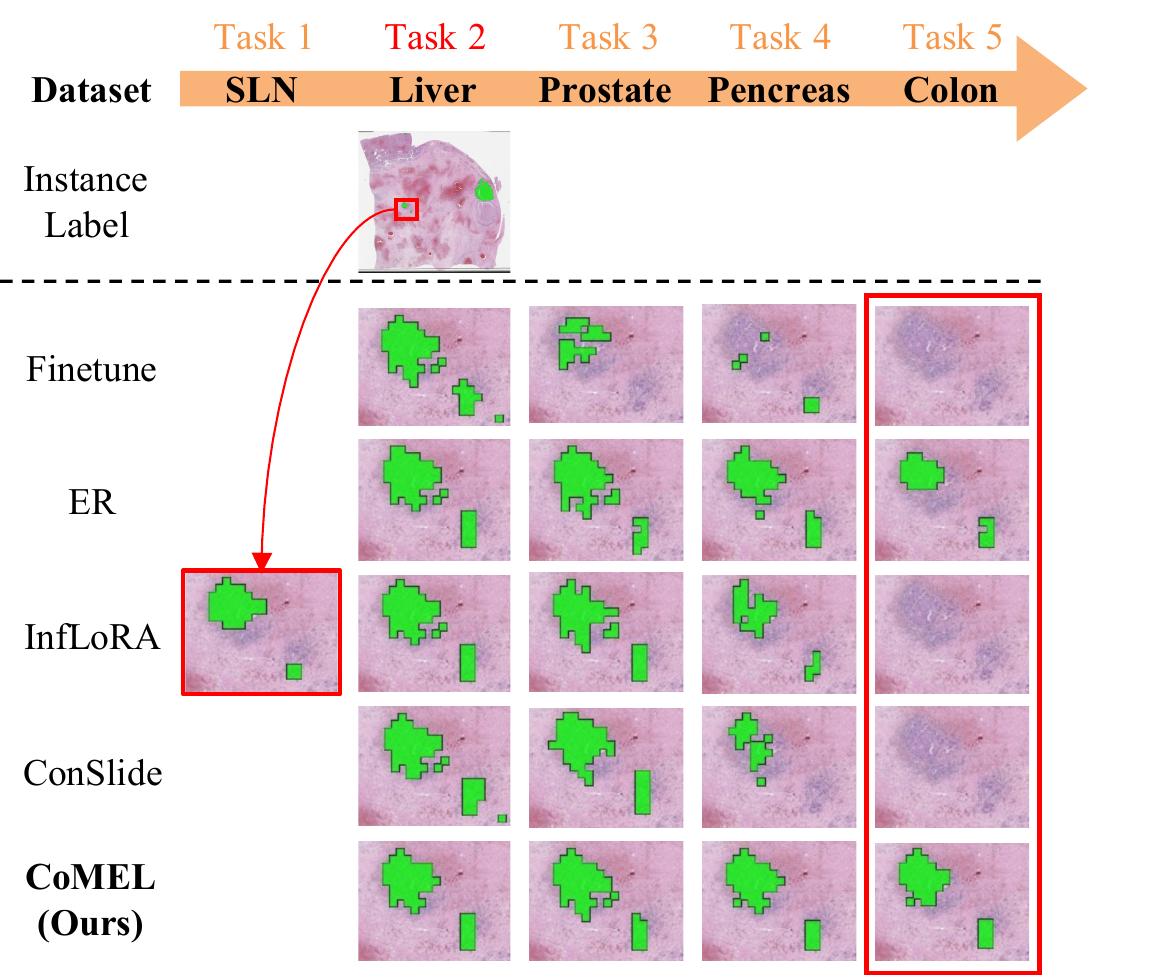}}
 \vskip -0.05in
\caption{Additional qualitative results of localization across sequential organ datasets under the continual MIL setup. Each column is the localization performance on Task 2 as the learned organ changes over sequential tasks. Each row corresponds to CL methods including CoMEL. CoMEL successfully preserved the localization quality across all tasks, while baselines increase false positives or false negatives.}
\label{fig:add_quali_instance_4}
\vskip -0.4in
\end{center}
\end{figure*}

We provide additional qualitative results for instance classification under continual MIL setup. From \cref{fig:add_quali_instance_1} to \cref{fig:add_quali_instance_4}, we can see that CoMEL can preserve the localized tumor region after training on all five organ datasets.

\section{Limitations}
In this work, we considered only a fixed sequence of disjoint tasks for the continual MIL setup. For example, the MIL models learn from datasets of distinct organs or subtypes for WSI analysis in our experiments. However, in real-world hospital settings, WSIs collected over a certain period include a mixture of various organs and tumor subtypes. In continual learning, such configuration has already been studied under the concept of blurry tasks \cite{kohonline}, where task boundaries are ambiguous. In this work, we did not consider such blurry tasks, leaving it as an interesting future work. Bag Prototypes-based Pseudo-Labeling (BPPL) module heavily relies on the accuracy of attention scores as pseudo-labels. Our ablation studies on BPPL hyperparameters in \cref{tab:ablation_bppl_supple_1} and \cref{tab:ablation_bppl_supple_2} indicate that localization performance is highly sensitive to hyperparameter selection which influences pseudo-label quality. The Orthogonal Weighted Low-Rank Adaptation (OWLoRA) effectively mitigated catastrophic forgetting in previous tasks. However, the basis of new tasks cannot be introduced indefinitely, as the maximum rank of a matrix is upper-bounded by its dimension. Therefore, OWLoRA has inherent limitations in learning an infinite sequence of sequential tasks.

\end{document}